%% file: main_en.tex
\documentclass[10pt]{article}
\usepackage[preprint]{tmlr}

\usepackage{amsmath,amssymb,amsthm,mathtools,bm,mathrsfs}
\usepackage{booktabs,tabularx,array,multirow}
\usepackage{enumitem}
\usepackage{placeins}
\usepackage{graphicx}
\usepackage{aliascnt}
\usepackage{xcolor}
\usepackage{algorithm}
\usepackage{algpseudocodex}
\usepackage{hyperref}
\definecolor{mydarkblue}{rgb}{0,0.08,0.45}
\hypersetup{colorlinks=true,linkcolor=mydarkblue,citecolor=mydarkblue,filecolor=mydarkblue,urlcolor=mydarkblue}
\usepackage{url}
\usepackage{etoolbox}
\AtBeginEnvironment{table}{\setlength{\abovecaptionskip}{0pt}\setlength{\belowcaptionskip}{8pt}}
\AtBeginEnvironment{table*}{\setlength{\abovecaptionskip}{0pt}\setlength{\belowcaptionskip}{8pt}}
\newcommand{\twoline}[2]{\begin{tabular}[b]{@{}c@{}}#1\\#2\end{tabular}}%
\usepackage[noabbrev,capitalise]{cleveref}
\graphicspath{{figures/}}

\input{preamble_main_en.tex}
\input{appendix_macros_en.tex}
\def\eqref#1{equation~\ref{#1}}

\title{Object-Flow Machine Learning: Operation-Level Surrogate Modeling for Self-Driving Laboratories}

\author{\name Yusuke Ozaki$^{1,2,3}$, Kazunari Kaizu$^{1,2,4}$, and Koichi Takahashi$^{1,2}$ \email yusuke.ozaki@riken.jp\\
      \addr $^1$RIKEN Center for Biosystems Dynamics Research\\
      $^2$Advanced General Intelligence for Science Program (AGIS)\\
      $^3$Kwansei Gakuin University, Department of Engineering, Program of Computer Science\\
      $^4$Cell Modeling and Simulation Group, The Exploratory Research Center on Life and Living Systems,\\
      National Institutes of Natural Sciences}

\begin{document}
\maketitle

\input{body_en.tex}
\bibliography{references}
\bibliographystyle{tmlr}

\input{appendix_en.tex}\end{document}

%% file: preamble_main_en.tex
\newcolumntype{Y}{>{\raggedright\arraybackslash}X}
\newcolumntype{C}{>{\centering\arraybackslash}X}

\newtheorem{theorem}{Theorem}[section]
\newaliascnt{assumption}{theorem}
\newtheorem{assumption}[assumption]{Assumption}
\aliascntresetthe{assumption}
\newaliascnt{proposition}{theorem}

\aliascntresetthe{proposition}
\newaliascnt{corollary}{theorem}

\aliascntresetthe{corollary}
\theoremstyle{remark}
\newaliascnt{remark}{theorem}
\newtheorem{remark}[remark]{Remark}
\aliascntresetthe{remark}
\crefname{theorem}{Theorem}{Theorems}
\crefname{assumption}{Assumption}{Assumptions}
\crefname{proposition}{Proposition}{Propositions}
\crefname{corollary}{Corollary}{Corollaries}
\crefname{remark}{Remark}{Remarks}
\crefname{algorithm}{Algorithm}{Algorithms}
\algrenewcommand\algorithmicrequire{\textbf{Input:}}
\algrenewcommand\algorithmicensure{\textbf{Output:}}

\newcommand{\Pp}{\mathbb P}
\newcommand{\E}{\mathbb E}
\newcommand{\R}{\mathbb R}
\newcommand{\Law}{\operatorname{Law}}
\newcommand{\Var}{\operatorname{Var}}
\newcommand{\Cov}{\operatorname{Cov}}

\newcommand{\norm}[1]{\left\lVert #1\right\rVert}
\newcommand{\abs}[1]{\left\lvert #1\right\rvert}

\newcommand{\DeltaIVR}{\Delta^{\mathrm{IVR}}}

\newcommand{\Pcal}{\mathcal P}
\newcommand{\Xcal}{\mathcal X}

\newcommand{\NRMSE}{\operatorname{NRMSE}}
\newcommand{\AULC}{\operatorname{AULC}}

\newcommand{\pa}{\operatorname{pa}}
\newcommand{\Xtgt}{\Xcal_{\mathrm{tgt}}}
\newcommand{\Ptgt}{\Pcal_{\mathrm{tgt}}}
\newcommand{\Psrc}{\Pcal_{\mathrm{src}}}

%% file: appendix_macros_en.tex
{\theoremstyle{plain}\newtheorem{stheorem}{Theorem}[section]}
\newaliascnt{sassumption}{stheorem}
{\theoremstyle{plain}\newtheorem{sassumption}[sassumption]{Assumption}}
\aliascntresetthe{sassumption}
\newaliascnt{slemma}{stheorem}
{\theoremstyle{plain}\newtheorem{slemma}[slemma]{Lemma}}
\aliascntresetthe{slemma}
\newaliascnt{sproposition}{stheorem}
{\theoremstyle{plain}\newtheorem{sproposition}[sproposition]{Proposition}}
\aliascntresetthe{sproposition}
\newaliascnt{sremark}{stheorem}
{\theoremstyle{remark}}
\aliascntresetthe{sremark}
\newcommand{\restatedname}{}

\crefname{sassumption}{Assumption}{Assumptions}
\crefname{slemma}{Lemma}{Lemmas}
\crefname{sremark}{Remark}{Remarks}
\crefname{stheorem}{Theorem}{Theorems}
\crefname{sproposition}{Proposition}{Propositions}

\providecommand{\F}{\mathcal F}
\providecommand{\E}{\mathbb E}
\providecommand{\Pp}{\mathbb P}
\providecommand{\R}{\mathbb R}
\providecommand{\Var}{\operatorname{Var}}
\providecommand{\Cov}{\operatorname{Cov}}
\providecommand{\Law}{\operatorname{Law}}

\providecommand{\norm}[1]{\left\lVert#1\right\rVert}
\providecommand{\abs}[1]{\left\lvert#1\right\rvert}

%% file: body_en.tex
\begin{abstract}
Self-driving laboratories change their experimental protocols as research progresses.
A surrogate model must then predict the outcomes of a new protocol from the measurements of earlier protocols.
Experimental protocols usually share common operations, and many experiments yield only terminal measurements.
Conventional surrogate models treat each protocol as a separate function, so they cannot reuse knowledge at the operation level.
In addition, existing surrogate models that use the structure of a process rely on observations of intermediate outputs.
We propose Object-Flow Machine Learning (OFML), a framework for operation-level surrogate models that share modules across protocols.
OFML represents each protocol as a typed directed acyclic graph whose nodes are operations.
OFML composes the modules along the graph to predict the outcomes of a protocol.
Each operation type has one module, so protocols that contain the same operation type reuse the same module.
OFML learns these shared modules only from terminal measurements.
The learned modules enable zero-shot transfer to new protocols and adaptation.
Under the assumptions that the surrogate model is correctly specified and that regularity conditions hold, we prove that the posterior mean converges uniformly to the true responses of the target protocols when integrated variance reduction (IVR) selects the observations.
In numerical experiments, OFML reduces the zero-shot error and speeds up adaptation compared with a surrogate model that does not share parameters across protocols.
Furthermore, with a correctly specified OFML surrogate model, the maximum posterior variance and the maximum error of the posterior mean decrease as IVR adds observations.
These results suggest that operations are a useful unit for reusing knowledge across protocols.
\end{abstract}

\section{Introduction}
Recent advances in the integration of AI and robotics are making closed-loop autonomous experimentation increasingly feasible, with hypothesis generation, experiment execution, observation, and model updating integrated into a single cycle
\citep{tom2024sdl,burger2020mobile,smith2026gpt5lab}.
In the life sciences and experimental chemistry, researchers do not simply explore a fixed experimental space in a single pass. Instead, in response to observations, they iteratively modify the experimental protocol itself, including reagents, measurement conditions, and the order of operations \citep{carbonell2018dbtl,opgenorth2019dbtl}.
Consequently, surrogate models for closed-loop autonomous experimentation require not only predictive accuracy within a single protocol but also the ability to reuse knowledge across different protocols. In addition, because real experiments are expensive, they require sample efficiency.

Black-box surrogate models that treat protocols independently cannot fully meet these requirements.
Such models are convenient for optimization with limited data \citep{rasmussen2006gp,shahriari2016bo,srinivas2010gpucb,kandasamy2015addgpucb,shields2021reactionbo}. However, they treat protocols as identifiers or flat feature vectors, so they discard structural information such as the order of operations, shared operations, and differences among measurement instruments \citep{garrido2020categorical,shields2021reactionbo}.
These limitations show the need for surrogate models that preserve protocol structure.

An additional challenge is that experimental protocols often provide only partial observations or terminal measurements: intermediate object states and operation-level effects remain unobserved. In the life sciences, for example, conventional single-cell RNA sequencing requires cell lysis, preventing repeated measurements of the same cell \citep{macosko2015dropseq,chen2022liveseq}. Consequently, a compositional surrogate model must infer operation-level maps from protocol-level terminal measurements without module-level supervision.

In this work, we propose Object-Flow Machine Learning (OFML) to address this problem. OFML represents each experimental protocol as a typed directed acyclic graph (DAG) whose nodes denote experimental operations and whose edges encode the flow of data and physical entities, such as samples, reagents, and containers. Each operation is modeled as a map from its explicit inputs to an updated state, so context relevant to its behavior should be represented in those inputs. OFML uses one learnable module for each operation type across protocols. The modules are trained jointly from protocol-level terminal losses. For a new target protocol, OFML composes the learned modules according to the graph of the target protocol to obtain an initial prediction. OFML then updates the shared modules as target observations arrive. Thus, OFML enables zero-shot transfer and adaptation across protocols while preserving their compositional structure.

We make three contributions:
\begin{enumerate}
\item
We formulate a function-learning problem over a set of typed protocols and propose OFML, which composes, along protocol DAGs, modules shared by
operation type.
\item
For adaptation, we prove uniform convergence of the posterior mean to the
true responses of the target protocols under IVR with noisy observations, assuming a correctly
specified Bayesian model and the regularity and sequential design conditions.
\item
Through numerical experiments, we show the empirical performance of OFML. We evaluate sharing based on protocol structure in zero-shot transfer and adaptation. We also evaluate the use of uncertainty in experiment selection. When the surrogate model is correctly specified, we confirm that the maximum posterior variance and the maximum error of the posterior mean, whose convergence to 0 is stated in \Cref{thm:compact-ofml}, decrease as observations are added.
\end{enumerate}

\begin{figure*}[t]
  \centering
  \includegraphics[width=\textwidth]{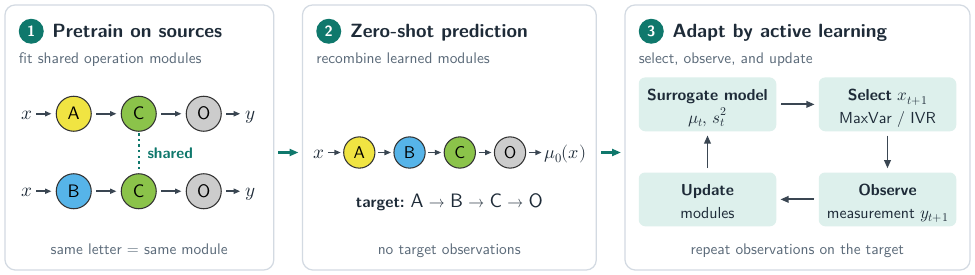}
  \caption{
  Overview of OFML. The modules of the operation types are trained jointly from terminal
  measurements on source protocols and recombined in the order of a previously
  unobserved target protocol to obtain predictions without target
  observations. After target deployment, points selected by MaxVar or IVR
  are observed sequentially, and their terminal measurements are used to update the shared modules.
  }
  \label{fig:method}
\end{figure*}

\section{Related Work}

\paragraph{Surrogate models that use structure or multiple tasks}
Multi-task Bayesian optimization shares observations across related tasks through a kernel over tasks, but it treats each task as an identifier \citep{Swersky2013}.
Cascade and multistage Bayesian optimization handle a single process with a fixed order of stages and observe the output of each stage \citep{Kusakawa2022,Torresi2025}.
Bayesian optimization of composite functions learns the expensive inner function of the objective from its observed outputs \citep{Astudillo2019}.
Bayesian optimization of function networks extends this approach to DAGs with a Gaussian process (GP) at each node and observes intermediate outputs \citep{Astudillo2021}, also with partial evaluations of the network \citep{Buathong2024}.
Neural architecture search predictors predict the performance of computational DAGs, but they feed the whole graph into a single kernel or network \citep{Ru2021,Wen2020}.
In contrast to these studies, OFML shares type-specific modules across a family of protocols and learns them only from terminal measurements.

\paragraph{Generalization by composing modules}
Studies that compose reusable modules to generalize to new problems either take the module combination as an input or search for it using data.
Neural module networks assemble modules from a question's syntax and train them jointly \citep{Andreas2016}.
Graph neural networks for DAGs also take the graph structure as input \citep{Thost2021}.
Modular meta-learning searches for a module combination that explains each task's data \citep{Alet2018}.
\citet{Schug2024} studied when hypernetworks can identify reusable modules from input--output pairs of tasks.
None of these studies places a posterior over the modules for sequential experimental design.
In OFML, the protocol itself determines the module combination.

\paragraph{Acquisition functions and consistency}
Acquisition functions for active learning use the posterior in different ways.
They select the point with the largest predictive variance, information gain, mutual information, or knowledge gradient \citep{MacKay1992,Houlsby2011,Krause2008,Frazier2009}.
We analyze IVR, which maximizes the expected decrease in the integrated variance \citep{Cohn1996}.
IVR is a stepwise uncertainty reduction (SUR) policy, which minimizes the expected residual uncertainty after each observation \citep{Bect2012}.
Broad classes of SUR policies are consistent for Gaussian-process posteriors \citep{Bect2019}, including those with vector outputs \citep{Stange2024}.
\Cref{thm:compact-ofml} shows the uniform convergence of the posterior mean under IVR for a correctly specified OFML surrogate model, whose posterior need not be Gaussian.

\paragraph{Deep ensembles and Bayesian inference}
Deep ensembles train the same network from several random initializations and give practical predictive uncertainty \citep{Lakshminarayanan2017}.
\citet{WilsonIzmailov2020} viewed them as an approximation to Bayesian model averaging, because the members fall into different basins of attraction of the posterior.
However, the members are not posterior samples, so the mean and variance of their predictions need not match those of the posterior predictive distribution \citep{Izmailov2021}.
\Cref{thm:compact-ofml} therefore does not cover OFML with a deep-ensemble approximation.
Moreover, the predictive uncertainty of deep ensembles degraded under dataset shift \citep{Ovadia2019}.

\section{Problem Setting}
\label{sec:problem-setting}

We consider a function-learning problem over a set of typed protocols.
Our objective is to reduce prediction error over all inputs of the target protocols.
\begin{figure*}[t]
  \centering
  \includegraphics[width=\textwidth]{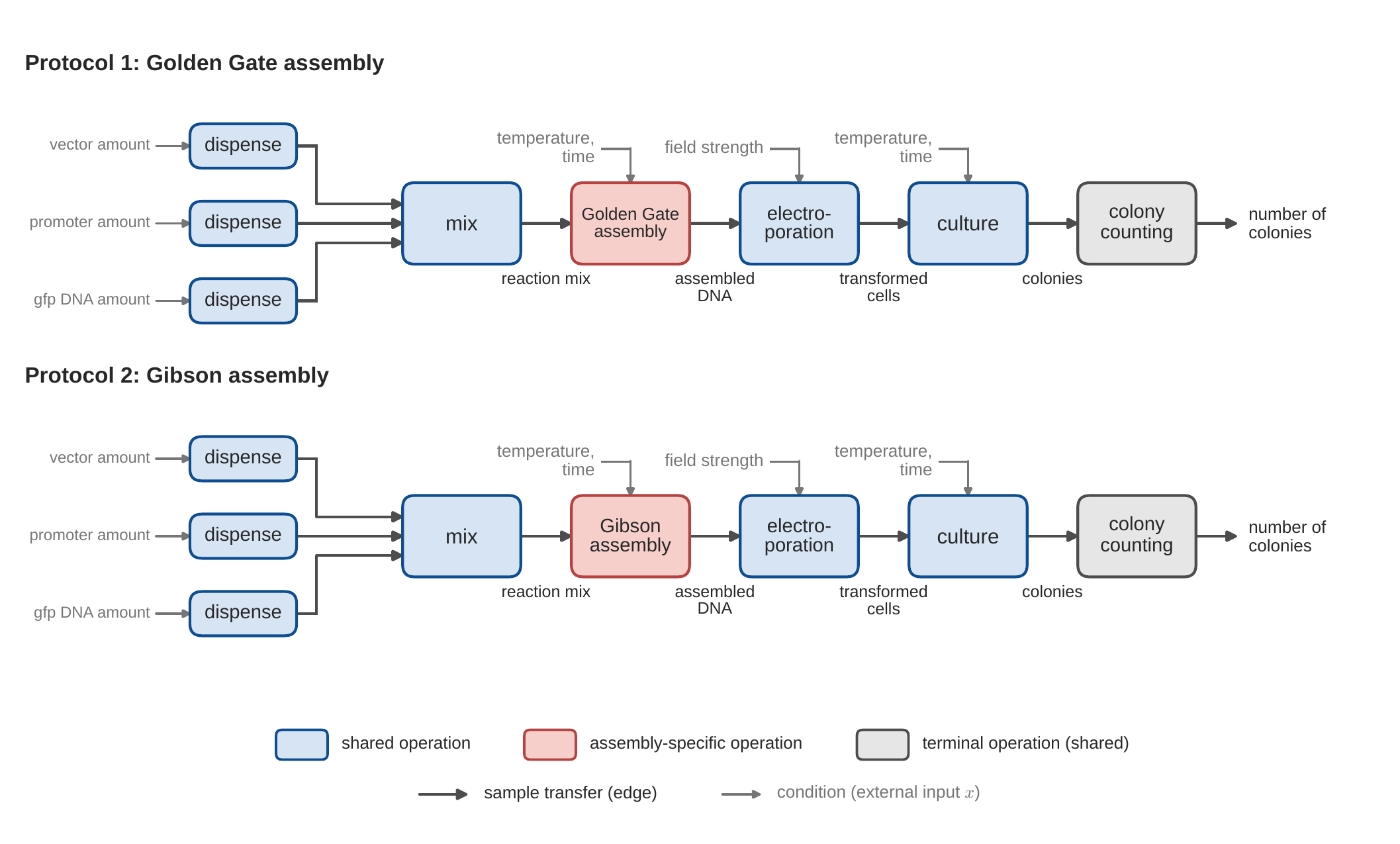}
  \caption{Example laboratory protocols. The two protocols are simplified versions of the DNA construction procedure of \citet{Shih2015}.
  They differ only in the operation type used for assembly. Each node groups several steps of that procedure. The mixing node takes the outputs of the dispensing nodes, from top to bottom, as its first, second, and third inputs.
  Only the terminal measurement is observed; the sample states written below the arrows are not observed.}
  \label{fig:lab-example}
\end{figure*}

Let $\mathcal T$ be a finite set of operation types and $\Pcal$ a set of protocols. Each protocol
$p\in\Pcal$ is a DAG with nodes $1,\ldots,J_p$ and terminal node $J_p$. The parent nodes of node $v$ are the nodes whose outputs node $v$ receives as inputs. Their set is $\pa_p(v)\subseteq\{1,\ldots,v-1\}$. Node $v$ performs an operation of type $\tau_{p,v}\in\mathcal T$. Let $n_{p,v}$ denote the external input dimension of node $v$. Node $v$ receives an external input $x_p^{(v)}\in\R^{n_{p,v}}$, such as an operating condition. For a node without external input,
$n_{p,v}=0$ and $x_p^{(v)}$ is the empty tuple. The design variable of $p$ collects the external inputs of all its nodes:
\[
x_p=(x_p^{(1)},\ldots,x_p^{(J_p)})\in\Xcal_p\subseteq\R^{n_{p,1}+\cdots+n_{p,J_p}}.
\]
The set of all inputs is the disjoint union $\Xcal:=\bigsqcup_{p\in\Pcal}\Xcal_p$.
Each element of $\Xcal$ specifies both a protocol $p$ and an input in $\Xcal_p$.

Each operation type $\tau$ has an input dimension $n^{\mathrm{in}}_\tau$ and an output dimension $n^{\mathrm{out}}_\tau$. Let $h=(h_\tau)_{\tau\in\mathcal T}$ be a tuple with one function $h_\tau:\R^{n^{\mathrm{in}}_\tau}\to\R^{n^{\mathrm{out}}_\tau}$ for each operation type $\tau$. Each $h_\tau$ maps the input of an operation of type $\tau$ to its output. For every protocol $p$ and every node $v$, we require
\[
n^{\mathrm{in}}_{\tau_{p,v}}=n_{p,v}+\sum_{w\in\pa_p(v)}n^{\mathrm{out}}_{\tau_{p,w}}.
\]
The terminal output is a scalar, that is, $n^{\mathrm{out}}_{\tau_{p,J_p}}=1$ for every $p$.
We define the input $u^h_v(x_p)$ of node $v$ recursively from its external input and the outputs of its parent nodes. We define the terminal function $f^h_p:\Xcal_p\to\R$ of protocol $p$ as the output of its terminal node:
\begin{equation}\label{eq:recursion}
u^h_v(x_p)
:=\Bigl(x_p^{(v)},\bigl(h_{\tau_{p,w}}(u^h_w(x_p))\bigr)_{w\in\pa_p(v)}\Bigr),
\qquad
f^h_p(x_p):=h_{\tau_{p,J_p}}\bigl(u^h_{J_p}(x_p)\bigr),
\end{equation}
where $w$ runs over $\pa_p(v)$ in increasing order, and each tuple is identified with the vector obtained by concatenating its components. Let $f^h$ be the function on $\Xcal$ that equals $f^h_p$ on each $\Xcal_p$ (\Cref{sec:notation}).

Let $f_p:\Xcal_p\to\R$ be the unknown true terminal response of protocol $p$.
Let $f$ be the function on $\Xcal$ that equals $f_p$ on each $\Xcal_p$.
We treat $f$ as a random function.
Executing $p$ at $x_p\in\Xcal_p$ yields only $y=f_p(x_p)+\varepsilon$, where $\varepsilon$ is the observation noise.
Outputs of nonterminal nodes are not observed.

\begin{assumption}[Shared Operations]
\label{ass:shared-operations}
The true function of each operation type is the same in every protocol. That is, there exists
$h^*=(h^*_\tau)_{\tau\in\mathcal T}$ such that, for every $p\in\Pcal$,
$f_p=f^{h^*}_p$ on $\Xcal_p$.
\end{assumption}

\paragraph{Example}
\Cref{fig:lab-example} shows two protocols that assemble DNA and introduce it into cells.
The set $\mathcal T$ consists of seven operation types: dispensing, mixing, Golden Gate assembly, Gibson assembly, electroporation, culture, and colony counting.
Each protocol has eight nodes.
Each dispensing node receives the amount of DNA as external input. The electroporation node receives the output of the assembly node and, as external input, the electric field strength.
Electroporation includes adding the prepared cells.
The colony-counting node is the terminal node and receives no external input.
Its output is the number of colonies. Only this output is observed.
For example, \Cref{ass:shared-operations} means that the true function of electroporation depends only on the state of the received sample and the field strength.
This function does not depend directly on which assembly operation produced the sample.

Let $\Psrc\subseteq\Pcal$ be the set of source protocols with observations.
Let $\Ptgt\subseteq\Pcal$ be the finite set of target protocols.
The set of inputs of the target protocols is $\Xtgt:=\bigsqcup_{p\in\Ptgt}\Xcal_p$.
Let $\mathcal D_0$ be the source data, a finite set of observations from the protocols in $\Psrc$.
We index time by the number $t=0,1,\ldots$ of additional observations.
At $t=0$, only the source data $\mathcal D_0$ are available.
A sequential policy is a rule that selects the next input based on the data obtained so far.
For each $t=1,2,\ldots$, the policy selects
$x_t\in\Xtgt$ based on $\mathcal D_{t-1}$.
We then observe $y_t=f(x_t)+\varepsilon_t$
and set $\mathcal D_t=\mathcal D_{t-1}\cup\{(x_t,y_t)\}$.
Let $\mathcal F_t:=\sigma(\mathcal D_0,x_1,y_1,\ldots,x_t,y_t)$ be the $\sigma$-algebra generated by the source data and the first $t$ observations.
$\mathcal F_t$ represents the information available at $t$.

\begin{assumption}[Observation Noise]
\label{ass:noise}
For each $t\ge1$, the conditional distribution of the noise $\varepsilon_t$ given $f$, $\mathcal F_{t-1}$, and $x_t$ is $\mathcal N(0,\xi^2)$, where $\xi^2>0$.
\end{assumption}

Let $\hat f_t$ be the predictor on $\Xtgt$ based on $\mathcal D_t$. For each $p\in\Ptgt$, let $\lambda_p$ be a probability measure on $\Xcal_p$.
This measure gives the weights of the inputs when we evaluate the prediction error for protocol $p$.
We give each target protocol equal weight and define the mean of the true response over all targets as
$\bar f_{\mathrm{tgt}}:=|\Ptgt|^{-1}\sum_{p\in\Ptgt}\int_{\Xcal_p}f_p\,\mathrm d\lambda_p$.
We define the root-mean-squared error (RMSE) of a predictor $g$ on $\Xtgt$.
We then measure the error at $t$ by the normalized RMSE (NRMSE):
\begin{equation}\label{eq:nrmse}
\operatorname{RMSE}(g):=\Bigl(\frac1{|\Ptgt|}\sum_{p\in\Ptgt}
\int_{\Xcal_p}\bigl(g-f_p\bigr)^2\,\mathrm d\lambda_p\Bigr)^{1/2},
\qquad
\NRMSE(t):=\frac{\operatorname{RMSE}(\hat f_t)}{\operatorname{RMSE}(\bar f_{\mathrm{tgt}})}.
\end{equation}
The denominator is the RMSE of the constant predictor $\bar f_{\mathrm{tgt}}$.
The learning curve shows $\NRMSE(i)$ against the number $i$ of additional observations.
For $t\ge1$, we summarize this curve up to $t$ additional observations by the area under the learning curve (AULC).
The AULC is the area under the piecewise-linear learning curve divided by $t$:
\begin{equation}\label{eq:aulc}
\AULC(t):=\frac1t\sum_{i=0}^{t-1}\frac{\NRMSE(i)+\NRMSE(i+1)}2.
\end{equation}
We aim to reduce $\NRMSE(t)$ for each $t\ge0$.
The prediction at $t=0$ is the zero-shot prediction, and we call $\NRMSE(0)$ the zero-shot error.
Updating the predictor with additional observations at $t\ge1$ is called adaptation.
With $T$ additional observations, we evaluate the final error of adaptation by $\NRMSE(T)$ and its speed by $\AULC(T)$.
\section{Object-Flow Machine Learning}
\label{sec:method}

Object-Flow Machine Learning (OFML) is a framework that constructs an operation-level surrogate model.
OFML learns a posterior distribution over the module parameters from terminal measurements alone.
\Cref{fig:method} summarizes learning, zero-shot transfer, and sequential adaptation.

\subsection{Operation-Level Surrogate Model}
\label{subsec:shared-operation-modules}

OFML assigns one parametric module to each operation type $\tau\in\mathcal T$.
For type $\tau$, $\theta_\tau$ denotes the module parameter, and $\Theta_\tau$ denotes its parameter space.
We write the module as $h_\tau(\cdot;\theta_\tau)$.
We write $\theta$ for the tuple of all module parameters and $\Theta$ for its parameter space:
\[
\theta=(\theta_\tau)_{\tau\in\mathcal T}\in\Theta,
\qquad
\Theta:=\prod_{\tau\in\mathcal T}\Theta_\tau.
\]
Each module maps $\R^{n^{\mathrm{in}}_\tau}$ to $\R^{n^{\mathrm{out}}_\tau}$.
Composing these modules according to the recursion in \eqref{eq:recursion} defines the surrogate model
\begin{equation}\label{eq:surrogate}
f_\theta:=f^{h(\cdot;\theta)}.
\end{equation}
Modules can be, for example, multilayer perceptrons, linear maps, or GP approximations based on a finite-dimensional feature map.

\begin{remark}
\label{rem:module-nonidentifiability}
Individual modules are in general not identified, that is, terminal measurements do not determine them uniquely. For example, suppose that every output of type $\tau$ enters one fixed input of type $\tau'$ and that this input receives nothing else. Applying an invertible map $\phi$ to the output of $\tau$ and $\phi^{-1}$ at that input leaves $f_\theta$ unchanged for all protocols, if the module families are closed under this change.
\end{remark}

\subsection{Learning and Prediction from Terminal Measurements}
\label{subsec:parameter-posterior}

Let $\gamma^2$ be the assumed noise variance. Given $\theta$, the surrogate model assumes $y\sim\mathcal N(f_\theta(x),\gamma^2)$ for each $(x,y)\in\mathcal D_t$, with noise independent of the earlier observations.
Up to a constant, the negative log-likelihood is
\[
\mathcal L_t(\theta):=\frac1{2\gamma^2}\sum_{(x,y)\in\mathcal D_t}\bigl(y-f_\theta(x)\bigr)^2.
\]
Because outputs of intermediate nodes are unobserved, $\mathcal L_t$ depends on $\theta$ only through
the composition $f_\theta$.

We place a prior $\Pi_{\mathrm{prior}}$ on $\theta$ and define the posterior at $t\ge0$ by
\begin{equation}\label{eq:posterior}
\Pi_t(\mathrm d\theta)\propto\exp\bigl(-\mathcal L_t(\theta)\bigr)\,\Pi_{\mathrm{prior}}(\mathrm d\theta).
\end{equation}
Because $\mathcal D_0$ contains the source data, $\Pi_0$ is the posterior given the source data.
We write $\boldsymbol\theta$ for the parameter regarded as a random variable and $\theta$ for a value of the parameter.
When the prior and likelihood match the data-generating process, $\Pi_t$ equals the conditional distribution
$\Law(\boldsymbol\theta\mid\mathcal F_t)$.
Through $\mathcal L_t$, an observation at $x\in\Xcal_p$ can change the posterior of $\theta_\tau$ for every type in $p$. The observation can also change the predictions for other protocols that share one or more operation types with $p$.
Therefore, source observations can be used to predict the terminal responses of a target protocol
before any observations from that target protocol are available.

Composition generally makes $f_\theta$ nonlinear in $\theta$. For this reason, $\Pi_t$ generally has no closed form.
We approximate $\Pi_t$ by $M$ parameter values $\theta^{(1)},\ldots,\theta^{(M)}$.
For Bayesian models, these parameter values are weighted particles that approximate $\Pi_t$. Each particle is a parameter value with a weight.
We obtain the particles by Monte Carlo methods such as sequential Monte Carlo \citep{DelMoral2006}.
For models trained by gradient-based optimization, such as neural networks, backpropagation computes the gradient of $\mathcal L_t$ with respect to $\theta_\tau$. Backpropagation passes the terminal errors backward along the protocol DAG and sums the gradients at nodes of the same type. We obtain the parameter values by approximation methods such as deep ensembles \citep{Lakshminarayanan2017}.
A deep ensemble trains the same network from several random initializations.
We use the resulting parameter values in place of samples from $\Pi_t$.

Suppose that $f_\theta(x)$ has a finite second moment under $\Pi_t$ for each $x\in\Xtgt$. Under $\Pi_t$, the random terminal function $f_{\boldsymbol\theta}$ restricted to $\Xtgt$ has the posterior mean $\mu_t$ and the posterior covariance $k_t$.
For $x,x'\in\Xtgt$, we define
\begin{equation}\label{eq:moments}
\mu_t(x):=\int_\Theta f_\theta(x)\,\Pi_t(\mathrm d\theta),
\qquad
k_t(x,x'):=\int_\Theta\bigl(f_\theta(x)-\mu_t(x)\bigr)\bigl(f_\theta(x')-\mu_t(x')\bigr)\,\Pi_t(\mathrm d\theta).
\end{equation}
We write the posterior variance as $s_t^2(x):=k_t(x,x)$.
The implementation replaces $\mu_t$ and $k_t$ by the mean and covariance computed from
$f_{\theta^{(1)}},\ldots,f_{\theta^{(M)}}$. \Cref{sec:gpm-gp,sec:gpm-nn} describe this computation for each surrogate model.
We use $\hat f_t:=\mu_t$ as the predictor on $\Xtgt$.
\section{Theoretical Analysis}
\label{sec:theory}

This section gives a result on adaptation.
Suppose that the OFML surrogate model is correctly specified and that IVR selects the observation points.
Then the posterior mean converges uniformly to the true responses of the target protocols.
This result concerns the terminal functions and does not require identifying individual modules.

\subsection{Uniform Consistency under IVR}
\label{subsec:target-consistency}

In this subsection, we assume that the surrogate model's noise variance $\gamma^2$ equals the true noise variance $\xi^2$.
We also assume that each $\Xcal_p$ is compact.
For a function $g$ on $\Xtgt$, we write its uniform norm as $\norm{g}_\infty:=\sup_{x\in\Xtgt}|g(x)|$.
In expressions involving the uniform norm, $f$ denotes its restriction to $\Xtgt$.

Fix a finite Borel measure $\nu$ on $\Xtgt$ to weight the posterior variance over the target inputs.
Assume that $\nu$ assigns positive measure to every nonempty open subset of $\Xtgt$.
We define the integrated posterior variance by $V_t:=\int_{\Xtgt}s_t^2(x')\,\nu(\mathrm dx')$.
The hypothetical observation $y^x$ represents a possible observation at a candidate point $x\in\Xtgt$.
We define its distribution by the current posterior $\Pi_t$ defined in \eqref{eq:posterior} and the surrogate model's likelihood:
\[
\boldsymbol\theta\mid\mathcal F_t\sim\Pi_t,
\qquad
y^x\mid\boldsymbol\theta,\mathcal F_t\sim\mathcal N\bigl(f_{\boldsymbol\theta}(x),\xi^2\bigr).
\]
Using $y^x$, we define the integrated posterior variance $V_{t+1}^x$ after observing $y^x$ and the IVR gain $\DeltaIVR_t(x)$:
\begin{equation}\label{eq:ivr-gain}
V_{t+1}^x:=\int_{\Xtgt}\Var\bigl(f_{\boldsymbol\theta}(x')\mid\mathcal F_t,y^x\bigr)\,\nu(\mathrm dx'),
\qquad
\DeltaIVR_t(x):=V_t-\E\bigl[V_{t+1}^x\mid\mathcal F_t\bigr].
\end{equation}
This gain is the expected variance reduction criterion of active learning \citep{Cohn1996},
defined for the integrated posterior variance of $f_{\boldsymbol\theta}$ on $\Xtgt$.

\begin{assumption}[Approximate IVR Maximization]
\label{ass:cx-design}
For each $t\ge0$, let $\eta_t\ge0$ be an $\mathcal F_t$-measurable tolerance for the maximization error.
Assume that $\eta_t\to0$ almost surely.
For each $t\ge0$, the point $x_{t+1}\in\Xtgt$ is $\mathcal F_t$-measurable and satisfies
\begin{equation}\label{eq:ivr-design}
\DeltaIVR_t(x_{t+1})\ge\sup_{x\in\Xtgt}\DeltaIVR_t(x)-\eta_t.
\end{equation}
Moreover, $\DeltaIVR_t$ is, almost surely, upper semicontinuous in $x$.
\end{assumption}

\begin{assumption}[Correct Specification]
\label{ass:correct-spec}
Almost surely, $f=f_{\boldsymbol\theta}$ on the inputs $\Xcal$ of all protocols for a random parameter $\boldsymbol\theta$ whose distribution equals the prior $\Pi_{\mathrm{prior}}$ of the surrogate model.
For each $t\ge0$, $\Pi_t$ is a version of the conditional distribution $\Law(\boldsymbol\theta\mid\mathcal F_t)$.
\end{assumption}

\begin{assumption}[Regularity]
\label{ass:regularity}
The true response satisfies $\E\norm{f}_\infty^2<\infty$.
For each $\theta$, each module $h_\tau(\cdot;\theta_\tau)$ is Lipschitz continuous with a Lipschitz constant $L_\tau(\theta)$.
Almost surely, the products of $L_\tau(\boldsymbol\theta)$ along every path in the target protocols are bounded above by a random variable $\Lambda$ with $\E[\Lambda^2]<\infty$.
\end{assumption}

\begin{assumption}[Measurability]
\label{ass:ofml-measurability}
The map $\theta\mapsto f_\theta|_{\Xtgt}$ from $\Theta$ to the space of continuous functions on $\Xtgt$ with the uniform norm is Borel measurable.
For each $t\ge0$, $\DeltaIVR_t(x)$ is jointly measurable in the sample point and $x\in\Xtgt$.
\end{assumption}

\begin{theorem}[Uniform Consistency of OFML]
\label{thm:compact-ofml}
Under \Cref{ass:noise} and \Cref{ass:cx-design,ass:correct-spec,ass:regularity,ass:ofml-measurability}, almost surely,
\[
\norm{s_t^2}_\infty\to0,
\qquad
\norm{\mu_t-f}_\infty\to0.
\]
\end{theorem}

\paragraph{Proof outline}
\Cref{thm:compact-ofml} follows from the more general \Cref{thm:general}.
We outline the proof of \Cref{thm:general} in four steps.
First, because $V_t$ is a nonnegative supermartingale, \Cref{ass:cx-design} implies that the maximum IVR gain converges to zero (\Cref{lem:ivr-drift}).
Second, the posterior mean after observing $y^x$ has mean squared prediction error no greater than that of any affine function of $y^x$.
Hence, the IVR gain is bounded below by a weighted integral of squared posterior covariances (\Cref{lem:noisy-ivr-bound}).
Third, combining these two steps with the modulus of continuity of $f$, we obtain $\norm{s_t^2}_\infty\to0$ (\Cref{lem:uniform-extinction}).
Fourth, by the modulus of continuity and L\'evy's upward theorem \citep{williams1991probability}, $\mu_t$ converges uniformly (\Cref{lem:uniform-moments}).
Because the third step shows that $s_t^2$ converges to zero, the limit of $\mu_t$ is $f$.

This result extends the consistency of SUR policies for Gaussian-process posteriors
\citep{Bect2019,Stange2024} to posteriors that arise from composing modules and need not be Gaussian.
The first step uses the supermartingale argument of these works, and the third and fourth steps use the modulus of continuity of $f$ and L\'evy's upward theorem in place of Gaussianity.

\begin{remark}
\label{rem:approx-ivr}
An implementation can maximize an acquisition function approximating the exact IVR gain defined in \eqref{eq:ivr-gain}.
Suppose that the maximization error converges to zero almost surely.
Suppose also that the uniform norm of the difference between this acquisition function and the IVR gain converges to zero almost surely.
Then the selected points satisfy \eqref{eq:ivr-design} in \Cref{ass:cx-design} with $\eta_t\to0$ almost surely (\Cref{sec:approx}).
Hence, the conclusions of \Cref{thm:compact-ofml} also hold when this acquisition function selects the observations.
\Cref{sec:experiments} uses the lower bound in \eqref{eq:lower-ivr} instead, and \Cref{prop:lower-bound} shows that the conclusions of \Cref{thm:compact-ofml} also hold for this choice.
\end{remark}

\section{Numerical Experiments}
\label{sec:experiments}

Hypothesis (i) is that composing modules and sharing parameters each reduce the zero-shot error and speed up adaptation.
Hypothesis (ii) concerns a correctly specified surrogate model. It is that uncertainty-based acquisition functions reduce the error faster than Space-filling and Random, and that the two quantities in \Cref{thm:compact-ofml} decrease as the number of observations increases.
\Cref{subsec:exp-transfer,subsec:exp-acquisition} test hypotheses (i) and (ii), respectively.
\Cref{subsec:exp-acquisition} also examines posteriors approximated by a deep ensemble.

\subsection{Common Setup}
\label{subsec:exp-metrics}

\paragraph{True functions and inputs}
The true function of each type was drawn from a zero-mean GP with a squared exponential kernel of signal standard deviation $0.70$.
The length-scale is $1.0$ for first-layer types and $0.5$ for the others.
We call the set of the true functions of all types one true function.
We set $\gamma^2=\xi^2=0.01$, as assumed in \Cref{subsec:target-consistency}.
Each protocol's external input $x$ was chosen from a grid of 64 equally spaced points on $[0,4]$, including both endpoints.
We take the measure $\lambda_p$ used in \eqref{eq:nrmse} to be uniform on this grid.

\paragraph{Surrogate models}
OFML-GPN is OFML in which the prior of the function of each type is the GP from which the true function was drawn. Therefore, the OFML-GPN surrogate model is correctly specified.
OFML-GPN approximates its posterior by sequential Monte Carlo with 16,384 particles \citep{DelMoral2006} (\Cref{sec:gpm-gp}).
OFML-Ens is OFML in which the function of each type is a multilayer perceptron.
OFML-Ens approximates its posterior by a deep ensemble of five members \citep{Lakshminarayanan2017}.
At each $t$, each member of a deep-ensemble surrogate model is retrained on $\mathcal D_t$ from its initial weights (\Cref{sec:gpm-nn}).

\paragraph{Computation of IVR}
We take the measure $\nu$ used in \eqref{eq:ivr-gain} to be uniform on the candidate points of $\Xtgt$ and define
\begin{equation}\label{eq:lower-ivr}
\underline\Delta_t(x):=\frac1{|\Xtgt|}\sum_{x'\in\Xtgt}\frac{k_t(x,x')^2}{s_t^2(x)+\xi^2}.
\end{equation}
IVR selects as $x_{t+1}$ the candidate point maximizing $\underline\Delta_t(x)$.
Here $k_t$ is the posterior covariance defined in \eqref{eq:moments}.
The quantity $\underline\Delta_t(x)$ is a lower bound on the IVR gain $\DeltaIVR_t(x)$ defined in \eqref{eq:ivr-gain} (\Cref{lem:noisy-ivr-bound}).
The conclusion of \Cref{thm:compact-ofml} also holds when observations are selected by maximizing $\underline\Delta_t$ (\Cref{rem:approx-ivr}).

\paragraph{Comparisons}
To compare two methods A and B, we computed the difference A minus B in NRMSE or AULC, defined in equations~\ref{eq:nrmse} and~\ref{eq:aulc}, for each true function.
We report the mean of this difference over the true functions and the 95\% percentile bootstrap interval of this mean from 10,000 resamplings of the true functions.
The 95\% coverage is the fraction of the grid points $x$ used for $\NRMSE(t)$ at which $f_p(x)$ lies
in the central 95\% interval of the posterior distribution of $f_p(x)$.

\Cref{sec:repro} describes the computing resources and software.

\subsection{Transfer Across Protocols}
\label{subsec:exp-transfer}

\begin{figure*}[t]
  \centering
  \includegraphics[width=\textwidth]{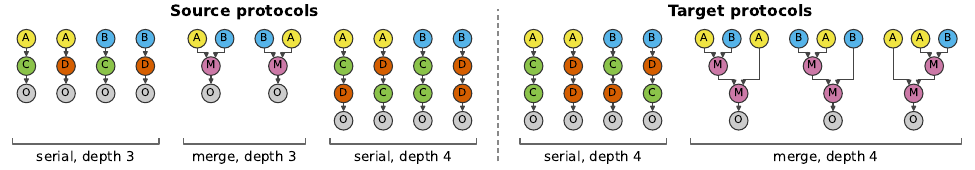}
  \caption{The 17 protocols of the benchmark in \Cref{subsec:exp-transfer}.
  An arrow points from a node to the node that receives its output.
  The first-layer types $\mathsf A$ and $\mathsf B$ receive the external input $x$.
  At a node of the merge type $\mathsf M$, the output of the left parent in the figure is the first input.}
  \label{fig:exp-protocols}
\end{figure*}

\paragraph{Data}
\Cref{fig:exp-protocols} shows the 10 source protocols and the 7 target protocols.
The input of the serial types $\mathsf{C}$ and $\mathsf{D}$ is one-dimensional, and that of the merge type $\mathsf M$ is two-dimensional.
In the serial target protocols, every input that a node receives lies within distance $0.5$, the length-scale of its type, of an input that a node of the same type receives in the source protocols.
In the merge target protocols, some inputs of $\mathsf M$ are farther than $0.5$ from every input of $\mathsf M$ in the source protocols.
The source data $\mathcal D_0$ consist of 8 observed points in each source protocol. \Cref{sec:gpm-bench} gives details.

\paragraph{Compared methods}
We compared OFML-GPN with five surrogate models chosen to separate the contributions of composition, parameter sharing, and type-specific modules.
\Cref{tab:exp-factors} summarizes the differences among the six surrogate models, and \Cref{sec:gpm-gp,sec:gpm-nn} give details.
Independent-GPN places a GP on the function of each node, as in function networks \citep{Astudillo2021}, and learns each target protocol independently from terminal measurements alone.
Shared-DAG-Ens and DAGNN-Cat-Ens feed a type embedding into a network shared across types.
Shared-Ens takes as input the external input $x$ and the number of times each type is used in the protocol.

\begin{table}[t]
  \centering
  \caption{Surrogate models compared in \Cref{subsec:exp-transfer}.
  The column Shared indicates whether the parameters are shared across protocols.
  The column Composed indicates whether the modules are composed along the DAG.
  The column Per type indicates whether each type has its own module.
  OFML-GPN and OFML-Ens are the surrogate models proposed in this work.
  DAGNN-Cat-Ens is DAGNN \citep{Thost2021} with its aggregation of the parent states replaced by concatenation.
  The other three are baselines constructed to separate the contributions of the components.}
  \label{tab:exp-factors}
  \small
  \setlength{\tabcolsep}{4pt}
  \begin{tabular}{@{}l ccc ll@{}}
    \toprule
    Surrogate model & Shared & Composed & Per type & Function model & Inference \\
    \midrule
    OFML-GPN & \checkmark & \checkmark & \checkmark & GP & Sequential Monte Carlo \\
    Independent-GPN & $\times$ & \checkmark & \checkmark & GP & Sequential Monte Carlo \\
    OFML-Ens & \checkmark & \checkmark & \checkmark & Multilayer perceptron & Deep ensemble \\
    Shared-DAG-Ens & \checkmark & \checkmark & $\times$ & Multilayer perceptron & Deep ensemble \\
    DAGNN-Cat-Ens & \checkmark & \checkmark & $\times$ & Gated recurrent unit & Deep ensemble \\
    Shared-Ens & \checkmark & $\times$ & $\times$ & Multilayer perceptron & Deep ensemble \\
    \bottomrule
  \end{tabular}
\end{table}

\paragraph{Procedure}
At $t=0$, all surrogate models except Independent-GPN predicted the target protocols from the source data $\mathcal D_0$ alone.
With each surrogate model, IVR selected $T=200$ observations, with all grid points of the 7 target protocols as candidate points.
We fixed the hyperparameters of the two GP surrogate models to their true values. For the four deep-ensemble surrogate models, we selected the hyperparameters by Bayesian optimization (\Cref{sec:gpm-hpo}).
Before seeing the results, we fixed a criterion that excludes from the comparison any true function on which the particle approximation of either GP surrogate model degenerates (\Cref{sec:gpm-exclusion}).
Of the 34 true functions drawn, 15 remained after the exclusion, and we compared the surrogate models on 12 of them selected by the replacement procedure in \Cref{sec:gpm-replacement}.
We report results on all 7 target protocols and separately on the serial and merge groups.
For each group, the denominator of $\NRMSE$ is the RMSE of the constant predictor given by the mean true response of that group.

\begin{figure*}[t]
  \centering
  \includegraphics[width=\textwidth]{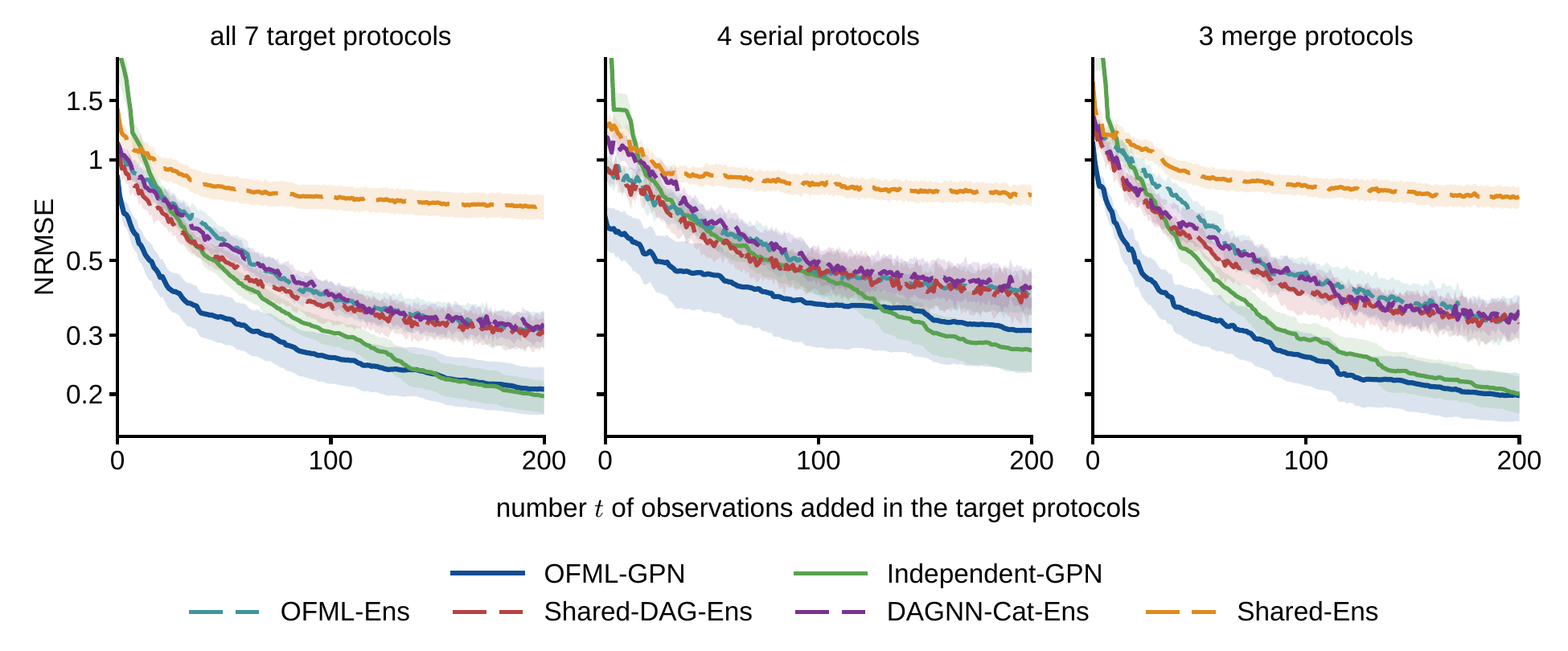}
  \caption{Comparison of operation-level surrogate models.
  Lines show the mean $\NRMSE(t)$ over the 12 true functions, and bands show the mean $\pm$ one standard error.
  Solid lines show the GP surrogate models, and dashed lines show the deep-ensemble surrogate models.
  Independent-GPN does not use the source data, so its prediction at $t=0$ is the prior mean 0.
  Its $\NRMSE(0)$ exceeds the upper limit of the vertical axis in all three panels (\Cref{tab:exp-transfer}).}
  \label{fig:exp-transfer}
\end{figure*}

\begin{table*}[t]
  \centering
  \caption{Metrics comparing the operation-level surrogate models. The values in the columns All, Serial, and Merge are computed on all 7 target
  protocols, the 4 serial ones, and the 3 merge ones, respectively; $T=200$.
  Values are means over the 12 true functions. Lower is better for all metrics.
  Bold marks the smallest value in each column.}
  \label{tab:exp-transfer}
  \footnotesize
  \setlength{\tabcolsep}{1.8pt}
  \input{generated_exp/table_obj1_v12_en.tex}
\end{table*}

\paragraph{Results}
\Cref{fig:exp-transfer} and \Cref{tab:exp-transfer} support hypothesis (i).
Sharing parameters reduced the zero-shot error.
The difference in $\NRMSE(0)$ between OFML-GPN and Independent-GPN over all target protocols was $-1.29$, with a 95\% interval of $[-2.14,-0.60]$.
Composing modules also reduced it.
The difference in $\NRMSE(0)$ between Shared-DAG-Ens and Shared-Ens was $-0.42$, with a 95\% interval of $[-0.70,-0.15]$ (\Cref{tab:gpm-paired}).
However, among the serial and merge groups, a zero-shot error below 1 was attained only in the serial group, and there only by OFML-GPN, OFML-Ens, and Shared-DAG-Ens.
In the merge group, about 27\% of the inputs of the third-layer $\mathsf M$ are farther than the length-scale $0.5$ from every input of $\mathsf M$ in the source protocols, on average over the 12 true functions (\Cref{sec:gpm-bench}).
In this group, $\NRMSE(0)$ was $1.11$ or larger for every surrogate model.

Sharing parameters sped up adaptation.
The difference in $\AULC(T)$ between OFML-GPN and Independent-GPN was $-0.12$, with a 95\% interval of $[-0.17,-0.07]$.
Composing modules also sped it up.
The difference in $\AULC(T)$ between Shared-DAG-Ens and Shared-Ens was $-0.37$, with a 95\% interval of $[-0.48,-0.26]$.
The difference in the final error $\NRMSE(T)$ between OFML-GPN and Independent-GPN was $+0.01$, with a 95\% interval of $[-0.03,+0.05]$ (\Cref{tab:gpm-paired,sec:gpm-diagnostics}).
In this benchmark, type-specific modules did not speed up adaptation.
OFML-Ens reduced the error more slowly than Shared-DAG-Ens, with a difference in $\AULC(T)$ of $+0.04$ and a 95\% interval of $[+0.02,+0.05]$.

\subsection{Acquisition Functions and Uniform Consistency}
\label{subsec:exp-acquisition}

\begin{figure*}[t]
  \centering
  \includegraphics[width=\textwidth]{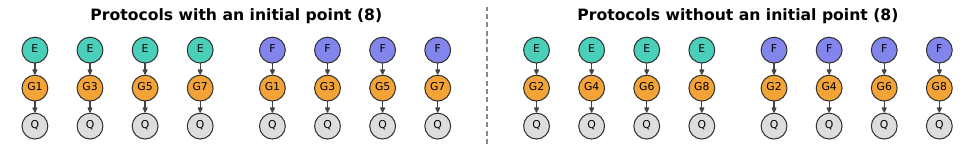}
  \caption{The 16 protocols of the benchmark in \Cref{subsec:exp-acquisition}, all of which are target protocols.
  Nodes and arrows are drawn as in \Cref{fig:exp-protocols}.
  The first-layer types $\mathsf E$ and $\mathsf F$ receive the external input $x$.
  The types $\mathsf{G1},\ldots,\mathsf{G8}$ are distinct from one another.}
  \label{fig:exp-acq-protocols}
\end{figure*}

\begin{figure*}[t]
  \centering
  \includegraphics[width=\textwidth]{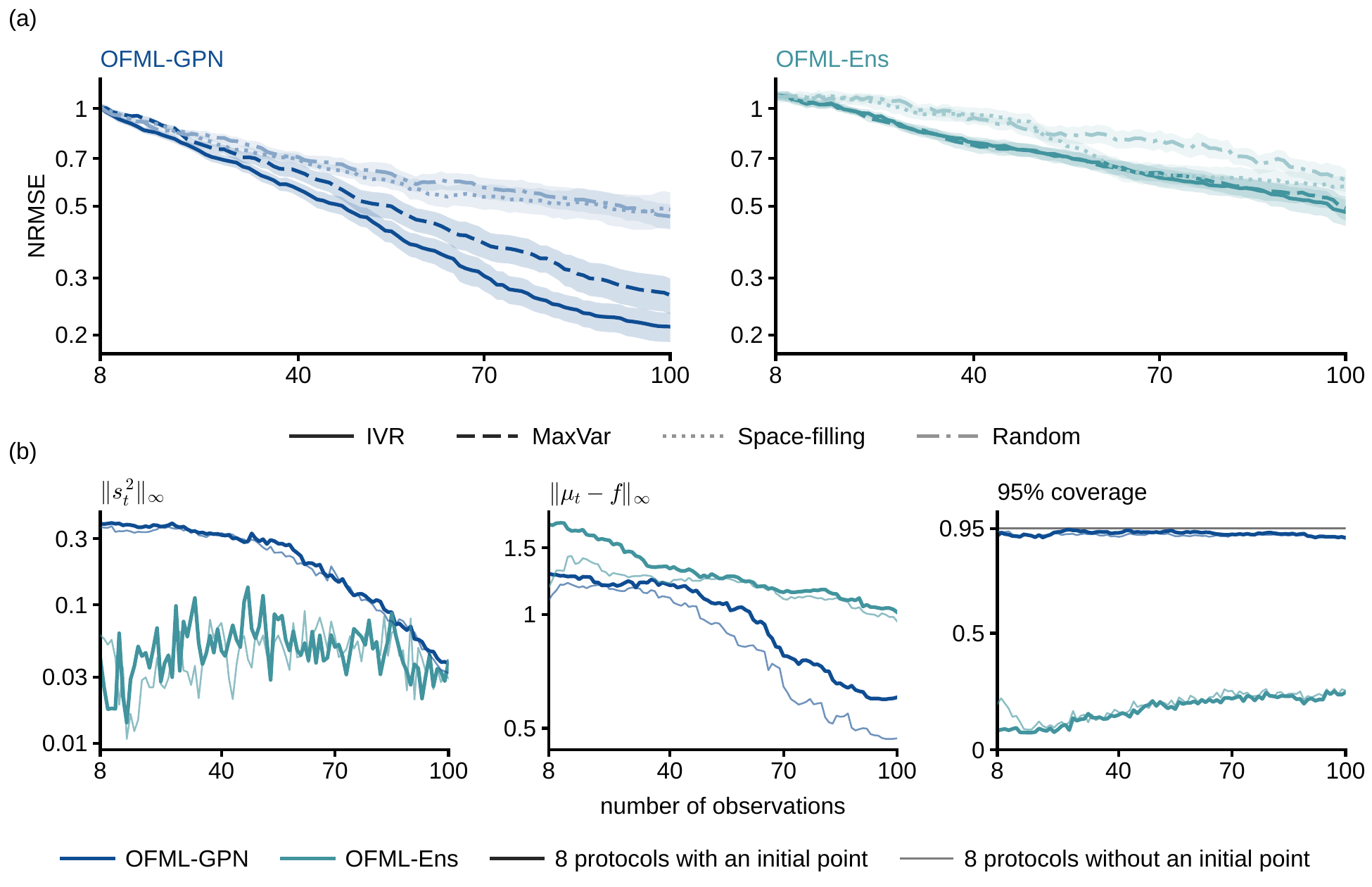}
  \caption{Comparison of acquisition functions (a), and trajectories of the two quantities of \Cref{thm:compact-ofml} and the 95\% coverage when IVR selects the observations (b).
  Values are means over the 16 true functions, and the bands in (a) show the mean $\pm$ one standard error.
  OFML-Ens uses the hyperparameters selected in \Cref{sec:acq-ens-selection}.
  In (b), the maxima and the coverage are computed on the grid points, separately for the protocols with an initial point and for the protocols without an initial point.
  The grey line marks the nominal coverage 0.95.
  For OFML-Ens, the posterior variance in (b) is computed with divisor 5.
  MaxVar and the coverage use this variance.
  IVR uses the covariance in \Cref{sec:gpm-nn}.}
  \label{fig:exp-acquisition}
\end{figure*}

\paragraph{Data}
\Cref{fig:exp-acq-protocols} shows the 16 target protocols. There are no source protocols.
The initial data $\mathcal D_0$ consist of one observed point in each of the 8 protocols on the left of the figure (\Cref{sec:acq-details}).

\paragraph{Compared methods}
We compared four acquisition functions.
IVR and MaxVar use predictive uncertainty, and Space-filling and Random do not.
MaxVar selects the candidate point with the largest $s_t^2(x)$.
Space-filling selects the candidate points in a fixed order (\Cref{sec:acq-details}).
Random selects a candidate point uniformly at random.
All four select only from the unobserved candidate points.
We examined the four acquisition functions with OFML-GPN, whose surrogate model is correctly specified, and with OFML-Ens, whose posterior is approximated by a deep ensemble.

\paragraph{Procedure}
For each combination of an acquisition function and a surrogate model, we made $T=92$ observations after the initial points.
All 1024 grid points of the 16 protocols were candidate points. There were 16 true functions.
We fixed the hyperparameters of OFML-GPN to their true values. For OFML-Ens, we selected the hyperparameters by Bayesian optimization on true functions separate from these 16 true functions (\Cref{sec:acq-ens-selection}).
In the runs with IVR, we recorded the two quantities in \Cref{thm:compact-ofml}, which are the maximum posterior variance $\norm{s_t^2}_\infty$ and the maximum error of the posterior mean $\norm{\mu_t-f}_\infty$.

\paragraph{Results}
\Cref{fig:exp-acquisition} supports hypothesis (ii) for OFML-GPN.
With OFML-GPN, the uncertainty-based acquisition functions reduced the error faster than Space-filling and Random.
The difference in the $\AULC(T)$ of OFML-GPN between IVR and Space-filling was $-0.17$, with a 95\% interval of $[-0.21,-0.14]$.
MaxVar showed the same tendency, and the differences of IVR and MaxVar from Random were larger in magnitude than those from Space-filling (\Cref{sec:acq-paired}).
With OFML-Ens, the uncertainty-based acquisition functions also reduced the error faster than Space-filling and Random (\Cref{sec:acq-paired}).
The difference in the $\AULC(T)$ of OFML-Ens between IVR and Space-filling was $-0.07$, with a 95\% interval of $[-0.13,-0.02]$.

With OFML-GPN and IVR, the two quantities in \Cref{thm:compact-ofml} decreased as the number of observations increased, even on the protocols without an initial point.
With OFML-Ens, the maximum error of the posterior mean stayed at or above $0.96$ even at 100 observations.
In addition, the maximum posterior variance of OFML-Ens remained small throughout.
\Cref{thm:compact-ofml} assumes a correctly specified surrogate model, and OFML-Ens is not one.
The 95\% coverage of OFML-Ens stayed between $0.07$ and $0.26$.
This coverage indicates that the posterior variance of OFML-Ens underestimated the magnitude of the prediction error (\Cref{sec:acq-uniform}).

\FloatBarrier
\section{Conclusion}
We consider a function-learning problem on a set of typed protocols and propose OFML.
OFML assigns one module to each operation type, shares the modules across protocols, and composes them along the protocol DAGs. OFML learns the modules from terminal measurements alone.
We prove that, if the surrogate model is correctly specified, the regularity conditions of \Cref{thm:compact-ofml} hold, and IVR selects the observations,
then the maximum posterior variance converges to 0 and the posterior mean converges uniformly to the true responses of the target protocols.

In the experiments, composing modules and sharing parameters each reduced the zero-shot error and sped up adaptation.
Furthermore, the zero-shot prediction of OFML was more accurate than the constant predictor given by the mean true response when every input of a node in the target protocols
lay within the length-scale of an input that nodes of the same type received in the source protocols.
These results suggest that operations are a useful unit for reusing knowledge across protocols.
For a correctly specified surrogate model, we also confirmed two results. First, uncertainty-based acquisition functions reduced the error faster than Space-filling and Random.
This also held for OFML-Ens, whose posterior is approximated by a deep ensemble.
Second, the maximum posterior variance and the maximum error of the posterior mean decreased as the number of observations increased.

We expect OFML to open various directions for future research. First, in the current implementation, the stability of the sequential Monte Carlo approximation to the posterior remains a limitation. The implementation already runs on GPUs, and its computation time grows linearly with the number of particles, so larger GPU resources allow more particles. Second, extending the problem setting to nonuniform observation costs and multiple observers is another direction, for which methods based on information gain per unit cost could be applied. Furthermore, connecting OFML to robots and implementing it stably with expressive neural networks will be key to applying OFML to real-world life science experiments. Realizing both is a first step toward the automation of life science.

%% file: generated_exp/table_obj1_v12_en.tex
\begin{tabular}{@{}l rrr rrr rrr@{}}
\toprule
 & \multicolumn{3}{c}{All} & \multicolumn{3}{c}{Serial} & \multicolumn{3}{c}{Merge}\\
\cmidrule(lr){2-4}\cmidrule(lr){5-7}\cmidrule(l){8-10}
Surrogate model & $\NRMSE(0)$ & $\NRMSE(T)$ & $\AULC(T)$ & $\NRMSE(0)$ & $\NRMSE(T)$ & $\AULC(T)$ & $\NRMSE(0)$ & $\NRMSE(T)$ & $\AULC(T)$\\
\midrule
OFML-GPN & \textbf{0.883} & 0.207 & \textbf{0.302} & \textbf{0.671} & 0.310 & \textbf{0.403} & \textbf{1.111} & \textbf{0.198} & \textbf{0.313} \\
Independent-GPN & 2.174 & \textbf{0.198} & 0.427 & 3.017 & \textbf{0.270} & 0.547 & 2.218 & 0.201 & 0.452 \\
OFML-Ens & 1.024 & 0.312 & 0.477 & 0.933 & 0.402 & 0.544 & 1.245 & 0.342 & 0.551 \\
Shared-DAG-Ens & 1.002 & 0.303 & 0.438 & 0.933 & 0.401 & 0.526 & 1.257 & 0.325 & 0.493 \\
DAGNN-Cat-Ens & 1.125 & 0.325 & 0.475 & 1.130 & 0.419 & 0.589 & 1.359 & 0.356 & 0.519 \\
Shared-Ens & 1.417 & 0.723 & 0.811 & 1.245 & 0.777 & 0.874 & 1.703 & 0.777 & 0.883 \\
\bottomrule
\end{tabular}

%% file: appendix_en.tex
\appendix
\crefalias{section}{appendix}
\crefalias{subsection}{subappendix}
\setcounter{figure}{0}\renewcommand{\thefigure}{S\arabic{figure}}
\setcounter{algorithm}{0}\renewcommand{\thealgorithm}{S\arabic{algorithm}}
\setcounter{table}{0}\renewcommand{\thetable}{S\arabic{table}}
\setcounter{equation}{0}\renewcommand{\theequation}{S\arabic{equation}}

\section*{Overview of the Appendix}
\Cref{sec:notation} fixes the notation.
\Cref{sec:main-proof,sec:ofml-proof,sec:approx} establish uniform consistency and extend it to the acquisition functions that implementations compute.
\Cref{sec:exp-gpmatched,sec:acq-details,sec:repro} provide experimental details, additional results, computing resources, and software information.

\section{Notation and Setup}\label{sec:notation}

Let $(\Omega,\F,\Pp)$ be the underlying probability space.
We use the filtration $\mathcal F_t$ and the target domain $\Xtgt$ defined in \Cref{sec:problem-setting}.
Write $\mathcal F_\infty:=\sigma\!\left(\bigcup_{t\ge0}\mathcal F_t\right)$.
For a topological space $E$, let $\mathcal B(E)$ denote its Borel $\sigma$-algebra.

A point of $\Xtgt$ is represented by $(p,z)$ with $z\in\Xcal_p$.
For $x=(p,z)$ and $x'=(q,z')$, define
\[
d(x,x')
:=
\begin{cases}
\min\{\norm{z-z'},1\},&p=q,\\
1,&p\ne q.
\end{cases}
\]
Here $\norm{\cdot}$ is the Euclidean norm.
On each $\Xcal_p$, the metric $d$ induces the same topology as the Euclidean metric.
Each $\Xcal_p$ is compact as in \Cref{subsec:target-consistency}, and $\Ptgt$ is finite.
Thus $(\Xtgt,d)$ is a compact metric space.

Let $C(\Xtgt)$ be the space of real-valued continuous functions with the uniform norm $\norm{\cdot}_\infty$ defined in \Cref{subsec:target-consistency}.
For $H\in C(\Xtgt^2)$, write $\norm{H}_\infty:=\sup_{(x,x')\in\Xtgt^2}\abs{H(x,x')}$.
Both spaces are separable Banach spaces and are standard Borel spaces with their Borel $\sigma$-algebras.
We regard $f$ as a $C(\Xtgt)$-valued random element.
Then $f$ denotes the restriction of the true response to $\Xtgt$.

Write $\Pi_t^f$ for the posterior distribution of $f_{\boldsymbol\theta}|_{\Xtgt}$ under $\Pi_t$.
This distribution is a probability measure on $C(\Xtgt)$.
The posterior mean and posterior covariance in \eqref{eq:moments} have the representations
\begin{align*}
\mu_t(x)
&=\int g(x)\,\Pi_t^f(\mathrm dg),\\
\mu_t^{(2)}(x,x')
&:=\int g(x)g(x')\,\Pi_t^f(\mathrm dg),\\
k_t(x,x')
&=\mu_t^{(2)}(x,x')-\mu_t(x)\mu_t(x'),\\
s_t^2(x)
&=k_t(x,x).
\end{align*}
Regular conditional distributions of $C(\Xtgt)$-valued random elements exist because $C(\Xtgt)$ is a standard Borel space.
We define the versions of the posterior moments by these integrals.

We use the finite Borel measure $\nu$, the integrated posterior variance $V_t$, and the equality $\gamma^2=\xi^2$ of \Cref{subsec:target-consistency}.
For each $t$ and candidate point $x\in\Xtgt$, define the hypothetical observation $y^x$ by
\[
g\mid\mathcal F_t\sim\Pi_t^f,
\qquad
y^x\mid g,\mathcal F_t\sim\mathcal N\bigl(g(x),\xi^2\bigr).
\]
The corresponding hypothetical update and IVR gain are
\[
V_{t+1}^x
:=
\int_{\Xtgt}
\Var\bigl(g(x')\mid\mathcal F_t,y^x\bigr)\,\nu(\mathrm dx'),
\qquad
\DeltaIVR_t(x)
:=
V_t-\E[V_{t+1}^x\mid\mathcal F_t].
\]
These definitions agree with \eqref{eq:ivr-gain}.

Suppose that $\Pi_t^f$ is a version of $\Law(f\mid\mathcal F_t)$.
We then realize the hypothetical observation with conditional distribution $y^x\mid f,\mathcal F_t\sim\mathcal N\bigl(f(x),\xi^2\bigr)$.
The pairs $(g,y^x)$ and $(f,y^x)$ have the same conditional distribution given $\mathcal F_t$.
All conditional expectations in the proofs use the true probability law and this conditional distribution for the hypothetical observation.

Suppose that $x_{t+1}$ is $\mathcal F_t$-measurable and that \Cref{ass:noise} holds.
The conditional distributions of $(f,y^{x_{t+1}})$ and $(f,y_{t+1})$ given $\mathcal F_t$ then coincide.
Consequently, when $\Pi_t^f$ is a version of $\Law(f\mid\mathcal F_t)$,
\begin{equation}
\E[V_{t+1}^{x_{t+1}}\mid\mathcal F_t]
=
\E[V_{t+1}\mid\mathcal F_t]
\quad\text{almost surely}.
\label{eq:selected-point-identity}
\end{equation}

\section{Proof of a General Version of \texorpdfstring{\Cref{thm:compact-ofml}}{Theorem 5.5}}\label{sec:main-proof}

\Cref{thm:general} below states the conclusions of \Cref{thm:compact-ofml} under conditions on the function-space posterior $\Pi_t^f$ alone.
\Cref{sec:ofml-proof} derives \Cref{thm:compact-ofml} by verifying these conditions for OFML.

\begin{sassumption}[Correct Posterior and Modulus]\label{main:assumption-5-1}
Let $f:\Omega\to C(\Xtgt)$ be a Borel measurable map with
$\E\norm{f}_\infty^2<\infty$.
For every $t$, assume that $\Pi_t^f$ is a version of $\Law(f\mid\mathcal F_t)$.
Let $\omega:[0,\infty)\to[0,\infty)$ be deterministic, continuous, and nondecreasing, with $\omega(0)=0$.
Let $L_f\ge0$ be a random variable with $\E[L_f^2]<\infty$.
Assume that, almost surely,
\[
\abs{f(x)-f(x')}
\le
L_f\,\omega\!\left(d(x,x')\right)
\qquad\forall x,x'\in\Xtgt.
\]
\end{sassumption}

\begin{sassumption}[Measurability of the IVR Gain]\label{ass:measurability}
For each $t\ge0$, $\DeltaIVR_t(x)$ is jointly measurable in the sample point and $x\in\Xtgt$.
\end{sassumption}

\begin{stheorem}\label{thm:general}
Under \Cref{main:assumption-5-1,ass:measurability,ass:noise,ass:cx-design},
\[
\norm{s_t^2}_\infty\longrightarrow0,
\qquad
\norm{\mu_t-f}_\infty\longrightarrow0
\quad\text{almost surely}.
\]
\end{stheorem}

\paragraph{Proof outline}
\Cref{subsec:target-consistency} outlines the proof of \Cref{thm:general} in four steps.
\Cref{lem:posterior-moments} provides the posterior moments and the conditional bound on the squared increments of $f$ used in the proof.
\Cref{lem:ivr-drift,lem:noisy-ivr-bound,lem:uniform-extinction} establish the first, second, and third steps, respectively.
\Cref{lem:uniform-moments} and the final proof of \Cref{thm:general} establish the fourth step.

\begin{slemma}
\label{lem:posterior-moments}
Under \Cref{main:assumption-5-1}, the integrals defining $\mu_t$ and $\mu_t^{(2)}$ are Bochner integrals in $C(\Xtgt)$ and $C(\Xtgt^2)$, respectively.
There exist versions such that, on one event of probability one common to all $t$ and all $x,x'\in\Xtgt$,
\begin{gather}
\mu_t(x)=\E[f(x)\mid\mathcal F_t],
\label{eq:evaluation-first}\\
\mu_t^{(2)}(x,x')=\E[f(x)f(x')\mid\mathcal F_t],
\label{eq:evaluation-second}\\
\E[(f(x')-f(x))^2\mid\mathcal F_t]\le A_t\omega(d(x',x))^2,
\label{eq:conditional-modulus}
\end{gather}
where $A_t:=\E[L_f^2\mid\mathcal F_t]$.
Moreover,
\[
A_\star:=\sup_{t\ge0}A_t<\infty,
\qquad
B_\star:=
\sup_{t\ge0}\E[\norm{f}_\infty^2\mid\mathcal F_t]<\infty
\quad\text{almost surely}.
\]
\end{slemma}

\begin{proof}
Because $\Xtgt$ is a compact metric space, $C(\Xtgt)$ and $C(\Xtgt^2)$ are separable.
By \Cref{main:assumption-5-1},
\[
\int\norm{g}_\infty^2\,\Pi_t^f(\mathrm dg)
=
\E[\norm{f}_\infty^2\mid\mathcal F_t]
<
\infty
\quad\text{almost surely}.
\]
Consequently, $g\mapsto g$ and $g\mapsto g\otimes g$ are Bochner integrable under $\Pi_t^f$.
Here $(g\otimes g)(x,x')=g(x)g(x')$.
Point evaluation is a continuous linear functional and therefore commutes with these integrals.
This gives \eqref{eq:evaluation-first} and \eqref{eq:evaluation-second}.

The kernel $\Pi_t^f$ is a version of $\Law(f\mid\mathcal F_t)$.
Integrating over each event in $\mathcal F_t$ identifies the Bochner integrals as the corresponding Banach-space-valued conditional expectations.
We use only scalar martingale convergence theorems below.

Let $Q$ be a countable dense subset of $\Xtgt$.
Square the modulus bound in \Cref{main:assumption-5-1} and take conditional expectations given $\mathcal F_t$.
Outside one null set, \eqref{eq:conditional-modulus} holds simultaneously for all $t$ and all $x,x'\in Q$.
For each sample point outside this null set, the posterior integral $\int(g(x')-g(x))^2\,\Pi_t^f(\mathrm dg)$ is continuous in $(x',x)$.
The dominated convergence theorem applies with dominator $4\norm{g}_\infty^2$.
Continuity therefore extends \eqref{eq:conditional-modulus} from $Q^2$ to $\Xtgt^2$.
The posterior integrals give versions of these conditional expectations on all of $\Xtgt^2$.

Finally, $(A_t)_t$ and $\bigl(\E[\norm{f}_\infty^2\mid\mathcal F_t]\bigr)_t$ are nonnegative integrable scalar martingales.
Each converges almost surely to a finite limit.
Each sequence is therefore bounded on an event of probability one.
Taking the intersection of these events proves the bounds on $A_\star$ and $B_\star$.
\end{proof}

\begin{slemma}\label{lem:ivr-drift}
Under \Cref{main:assumption-5-1,ass:noise}, if $x_{t+1}$ is $\mathcal F_t$-measurable for every $t$, then $(V_t)_t$ is a nonnegative integrable supermartingale and
\[
V_t-\E[V_{t+1}\mid\mathcal F_t]
=
\DeltaIVR_t(x_{t+1})
\ge0
\quad\text{almost surely}.
\]
Moreover,
\[
\sum_{t=0}^{\infty}\DeltaIVR_t(x_{t+1})<\infty
\quad\text{almost surely}.
\]
If \Cref{ass:measurability,ass:cx-design} also hold, then
\[
\sup_{x\in\Xtgt}\DeltaIVR_t(x)\longrightarrow0
\quad\text{almost surely}.
\]
\end{slemma}

\begin{proof}
For each $x'$, the conditional law of total variance gives
\[
s_t^2(x')
=
\E[s_{t+1}^2(x')\mid\mathcal F_t]
+
\Var\!\left(
\E[f(x')\mid\mathcal F_{t+1}]
\,\middle|\,\mathcal F_t
\right).
\]
The integrands are nonnegative.
The conditional Tonelli theorem therefore gives
\[
V_t-\E[V_{t+1}\mid\mathcal F_t]
=
\int_{\Xtgt}
\Var\!\left(
\E[f(x')\mid\mathcal F_{t+1}]
\,\middle|\,\mathcal F_t
\right)\nu(\mathrm dx')
\ge0.
\]
The left-hand side $V_t-\E[V_{t+1}\mid\mathcal F_t]$ is the expected one-step decrease of $(V_t)_t$ given $\mathcal F_t$.
By \eqref{eq:selected-point-identity}, this expected decrease equals $\DeltaIVR_t(x_{t+1})$.

We have $V_t\ge0$ and $\E V_t\le\nu(\Xtgt)\E\norm{f}_\infty^2<\infty$.
Hence $(V_t)_t$ is a nonnegative integrable supermartingale.
Telescoping expectations gives $\sum_{t=0}^{\infty}\E\DeltaIVR_t(x_{t+1})\le\E V_0$.
Tonelli's theorem implies $\sum_t\DeltaIVR_t(x_{t+1})<\infty$ almost surely.
In particular, $\DeltaIVR_t(x_{t+1})\to0$ almost surely.

The conditional law of total variance gives nonnegativity, \Cref{ass:cx-design} gives the upper bound, and the preceding convergence gives
\begin{align*}
0
&\le
\sup_{x\in\Xtgt}\DeltaIVR_t(x)\\
&\le
\DeltaIVR_t(x_{t+1})+\eta_t\\
&\longrightarrow0
\quad\text{almost surely}.
\end{align*}
This proves the final conclusion.
\end{proof}

\begin{slemma}\label{lem:noisy-ivr-bound}
Under \Cref{main:assumption-5-1}, for the hypothetical observation defined in \Cref{sec:notation}, almost surely, for every $t$ and every $x\in\Xtgt$,
\[
\DeltaIVR_t(x)
\ge
\int_{\Xtgt}
\frac{k_t(x,x')^2}{s_t^2(x)+\xi^2}\,\nu(\mathrm dx').
\]
\end{slemma}

\begin{proof}
Work on the event of probability one from \Cref{lem:posterior-moments} where $\int\norm{g}_\infty^2\,\Pi_t^f(\mathrm dg)<\infty$ for every $t$.
We work conditionally on $\mathcal F_t$.
The hypothetical observation satisfies $\E[y^x\mid f,\mathcal F_t]=f(x)$ and $\Var(y^x\mid f,\mathcal F_t)=\xi^2$.
Consequently, $\Cov(y^x,f(x')\mid\mathcal F_t)=k_t(x,x')$ and $\Var(y^x\mid\mathcal F_t)=s_t^2(x)+\xi^2$.
The latter variance is positive because $\xi^2>0$.

Conditionally on $\mathcal F_t$, the conditional expectation $\E[f(x')\mid\mathcal F_t,y^x]$ is the mean-square projection of $f(x')$ onto the square-integrable functions of $y^x$.
Affine functions of $y^x$ belong to this space.
Comparison with the optimal affine predictor and the variance identity give
\[
\begin{aligned}
\Var\!\left(
\E[f(x')\mid\mathcal F_t,y^x]
\,\middle|\,\mathcal F_t
\right)
&\ge
\frac{k_t(x,x')^2}{\Var(y^x\mid\mathcal F_t)}\\
&=
\frac{k_t(x,x')^2}{s_t^2(x)+\xi^2}.
\end{aligned}
\]
The conditional law of total variance for the hypothetical update gives
\[
\DeltaIVR_t(x)
=
\int_{\Xtgt}
\Var\!\left(
\E[f(x')\mid\mathcal F_t,y^x]
\,\middle|\,\mathcal F_t
\right)\nu(\mathrm dx').
\]
Integrating the preceding bound proves the claim.
\end{proof}

\begin{slemma}
\label{lem:uniform-extinction}
Under \Cref{main:assumption-5-1}, if $\sup_{x\in\Xtgt}\DeltaIVR_t(x)\to0$ almost surely, then
\[
\norm{s_t^2}_\infty\longrightarrow0
\quad\text{almost surely}.
\]
\end{slemma}

\begin{proof}
Work on an event of probability one where \Cref{lem:posterior-moments,lem:noisy-ivr-bound} hold and the maximum IVR gain converges to zero.
Because $\Pi_t^f$ is a version of $\Law(f\mid\mathcal F_t)$,
\[
s_t^2(x)
\le
\int g(x)^2\,\Pi_t^f(\mathrm dg)
\le
\E[\norm{f}_\infty^2\mid\mathcal F_t].
\]
Thus \Cref{lem:posterior-moments} gives $S_\star:=\sup_{t\ge0}\norm{s_t^2}_\infty\le B_\star<\infty$.

Suppose that the claimed convergence fails.
There exist $\underline s^2>0$, a subsequence $(t_j)_j$, and maximizers $\tilde x_j\in\Xtgt$ such that $s_{t_j}^2(\tilde x_j)=\norm{s_{t_j}^2}_\infty\ge\underline s^2$.
The maximizers exist because $s_t^2$ is continuous and $\Xtgt$ is compact.
By the covariance identity, conditional Cauchy--Schwarz, and \eqref{eq:conditional-modulus}, respectively,
\begin{align*}
\abs{k_t(x,x')-s_t^2(x)}
&=
\abs{\Cov(f(x),f(x')-f(x)\mid\mathcal F_t)}\\
&\le
\sqrt{s_t^2(x)}
\sqrt{\E[(f(x')-f(x))^2\mid\mathcal F_t]}\\
&\le
\sqrt{A_\star S_\star}\,\omega(d(x',x)).
\end{align*}
Choose $\rho>0$ such that $\sqrt{A_\star S_\star}\,\omega(\rho)\le\underline s^2/2$.
Such a choice exists because $\omega$ is continuous at zero and $\omega(0)=0$.
If $\sqrt{A_\star S_\star}=0$, every positive $\rho$ satisfies this bound.
For $d(x',\tilde x_j)<\rho$, we obtain $k_{t_j}(\tilde x_j,x')\ge\underline s^2/2$.

Compactness and the condition on $\nu$ imply
\[
\underline\nu(\rho)
:=
\inf_{x\in\Xtgt}
\nu\bigl(\{x'\in\Xtgt:d(x,x')<\rho\}\bigr)
>
0.
\]
Indeed, choose a finite set whose open balls of radius $\rho/2$ cover $\Xtgt$.
For every $x$, one center lies at distance less than $\rho/2$ from $x$.
The open ball of radius $\rho/2$ at that center lies inside the open ball of radius $\rho$ at $x$.
Each of the finitely many balls of radius $\rho/2$ has positive measure.
Their minimum measure therefore bounds $\underline\nu(\rho)$ from below.

The integral consequently satisfies
\begin{equation}
\int_{\Xtgt}
\frac{k_{t_j}(\tilde x_j,x')^2}
{s_{t_j}^2(\tilde x_j)+\xi^2}\,\nu(\mathrm dx')
\ge
\frac{\underline s^4\,\underline\nu(\rho)}
{4(S_\star+\xi^2)}
>
0.
\label{eq:uniform-extinction-integral}
\end{equation}
By \Cref{lem:noisy-ivr-bound}, the same positive constant bounds $\DeltaIVR_{t_j}(\tilde x_j)$ from below.
This contradicts $\sup_x\DeltaIVR_t(x)\to0$ and proves the claim.
\end{proof}

\begin{slemma}
\label{lem:uniform-moments}
Under \Cref{main:assumption-5-1}, there exist random elements $\mu_\infty\in C(\Xtgt)$ and $\mu_\infty^{(2)}\in C(\Xtgt^2)$ such that
\[
\norm{\mu_t-\mu_\infty}_\infty\longrightarrow0,
\qquad
\norm{\mu_t^{(2)}-\mu_\infty^{(2)}}_\infty\longrightarrow0
\quad\text{almost surely}.
\]
Moreover, $k_t$ and $s_t^2$ have uniform limits $k_\infty$ and $s_\infty^2$ satisfying
\[
\norm{k_t-k_\infty}_\infty\longrightarrow0,
\qquad
\norm{s_t^2-s_\infty^2}_\infty\longrightarrow0
\quad\text{almost surely}.
\]
\end{slemma}

\begin{proof}
On the common event of \Cref{lem:posterior-moments}, conditional Jensen and \eqref{eq:conditional-modulus} give $\abs{\mu_t(x')-\mu_t(x)}\le\sqrt{A_\star}\,\omega(d(x',x))$.
Write $B_t:=\E[\norm{f}_\infty^2\mid\mathcal F_t]$.
For $b_1,b_2,b'_1,b'_2\in\Xtgt$, decompose the product difference as
\[
\begin{aligned}
f(b'_1)f(b'_2)-f(b_1)f(b_2)
&=
\bigl(f(b'_1)-f(b_1)\bigr)f(b'_2)\\
&\quad+
f(b_1)\bigl(f(b'_2)-f(b_2)\bigr).
\end{aligned}
\]
Conditional Cauchy--Schwarz and the definitions of $A_\star$ and $B_\star$ give
\begin{align*}
&\abs{\mu_t^{(2)}(b'_1,b'_2)-\mu_t^{(2)}(b_1,b_2)}\\
&\quad\le
\sqrt{A_tB_t}
\bigl\{
\omega(d(b'_1,b_1))+\omega(d(b'_2,b_2))
\bigr\}\\
&\quad\le
\sqrt{A_\star B_\star}
\bigl\{
\omega(d(b'_1,b_1))+\omega(d(b'_2,b_2))
\bigr\}.
\end{align*}
Thus both sequences are uniformly equicontinuous.
For each sample point, these moduli are common to all $t$.

Let $Q$ be the countable dense set used in \Cref{lem:posterior-moments}.
For each $\bar x,\bar x'\in Q$, apply L\'evy's upward theorem to \eqref{eq:evaluation-first} and \eqref{eq:evaluation-second}.
It gives almost-sure convergence of $\mu_t(\bar x)$ and $\mu_t^{(2)}(\bar x,\bar x')$.
Take the intersection of the countably many events on which these convergences hold.
Choose finite nets with centers in $Q$ and $Q^2$ for the compact spaces $\Xtgt$ and $\Xtgt^2$.
Convergence at the centers and the common moduli show that both sequences are uniformly Cauchy.
Their uniform limits are continuous.
As limits in separable Banach spaces, these limits are random elements.
Subtracting $\mu_t(x)\mu_t(x')$ and restricting to the diagonal preserve uniform convergence.
This proves the assertions about $k_t$ and $s_t^2$.
\end{proof}

\begin{proof}[Proof of \Cref{thm:general}]
\Cref{lem:ivr-drift} gives $\sup_x\DeltaIVR_t(x)\to0$ almost surely.
\Cref{lem:uniform-extinction} therefore proves the first conclusion.

Let $Q$ be the countable dense set used above.
By \Cref{lem:uniform-moments}, the posterior moments have uniform limits.
For each $\bar x\in Q$, \Cref{lem:posterior-moments} and L\'evy's upward theorem identify these limits as
\[
\mu_\infty(\bar x)
=
\E[f(\bar x)\mid\mathcal F_\infty],
\qquad
\mu_\infty^{(2)}(\bar x,\bar x)
=
\E[f(\bar x)^2\mid\mathcal F_\infty].
\]
Uniform convergence of the posterior moments and uniform vanishing of the posterior variance give $0=s_\infty^2(\bar x)=\Var(f(\bar x)\mid\mathcal F_\infty)$.
Hence $f(\bar x)=\mu_\infty(\bar x)$ almost surely.
Take one common event of probability one for every $\bar x\in Q$.
Continuity of $f$ and $\mu_\infty$ extends this equality to all of $\Xtgt$.
Therefore,
\[
\norm{\mu_t-f}_\infty
=
\norm{\mu_t-\mu_\infty}_\infty
\longrightarrow0
\quad\text{almost surely}.
\]
This proves the second conclusion.
\end{proof}

\section{Proof of \texorpdfstring{\Cref{thm:compact-ofml}}{Theorem 5.5}}
\label{sec:ofml-proof}

The surrogate model $f_\theta=f^{h(\cdot;\theta)}$ is defined in \eqref{eq:surrogate}.
Write $u_v^\theta:=u_v^{h(\cdot;\theta)}$ for the input of node $v$ obtained from \eqref{eq:recursion}.
Its output at $x\in\Xcal_p$ is $h_{\tau_{p,v}}\bigl(u_v^\theta(x);\theta_{\tau_{p,v}}\bigr)$.

\Cref{ass:correct-spec,ass:ofml-measurability} provide the correct parameter posterior and the measurable map needed to obtain the function-space posterior.
\Cref{ass:regularity} supplies the integrability and modulus conditions in \Cref{main:assumption-5-1}.
The IVR gain condition in \Cref{ass:ofml-measurability} supplies \Cref{ass:measurability}.
\Cref{ass:noise,ass:cx-design} supply the remaining conditions of \Cref{thm:general}.

\begin{slemma}
\label{lem:ofml-conditional-pushforward}
For a Borel measurable map $\theta\mapsto f_\theta|_{\Xtgt}$ from $\Theta$ to $C(\Xtgt)$ and a $\Theta$-valued random element $\boldsymbol\theta$,
if $\Pi_t$ is a version of $\Law(\boldsymbol\theta\mid\mathcal F_t)$ and $f=f_{\boldsymbol\theta}|_{\Xtgt}$ almost surely, then the pushforward kernel
\[
\Pi_t^f(U)
:=
\Pi_t\!\left(
\{\theta\in\Theta:f_\theta|_{\Xtgt}\in U\}
\right),
\qquad
U\in\mathcal B\!\left(C(\Xtgt)\right),
\]
is a version of $\Law(f\mid\mathcal F_t)$.
\end{slemma}

\begin{proof}
By the conditional distribution of $\boldsymbol\theta$ and the almost-sure identity for $f$, for every $U\in\mathcal B(C(\Xtgt))$,
\[
\begin{aligned}
\Pi_t^f(U)
&=
\E\!\left[
\mathbf 1_{\{f_{\boldsymbol\theta}|_{\Xtgt}\in U\}}
\,\middle|\,\mathcal F_t
\right]\\
&=
\E\!\left[
\mathbf 1_{\{f\in U\}}
\,\middle|\,\mathcal F_t
\right]
\quad\text{almost surely}.
\end{aligned}
\]
The Borel measurability of $\theta\mapsto f_\theta|_{\Xtgt}$ makes $\Pi_t^f$ an $\mathcal F_t$-measurable probability kernel.
Thus $\Pi_t^f$ satisfies the definition of a regular conditional distribution of $f$ given $\mathcal F_t$.
\end{proof}

Fix $p\in\Ptgt$.
For $x\in\Xcal_p$, write $x^{(v)}$ for the external input of node $v$.
Choose deterministic constants $L^{\mathrm{in}}_{p,v}\ge0$ satisfying
\begin{equation}
\norm{x^{(v)}-x'^{(v)}}
\le
L^{\mathrm{in}}_{p,v}\,d(x,x')
\qquad\forall x,x'\in\Xcal_p.
\label{eq:ofml-input-composition-lipschitz}
\end{equation}
The choice $L^{\mathrm{in}}_{p,v}:=\max\{1,\operatorname{diam}(\Xcal_p)\}$ satisfies \eqref{eq:ofml-input-composition-lipschitz}.
Here $\operatorname{diam}(\Xcal_p)$ is the finite Euclidean diameter of $\Xcal_p$.

\begin{slemma}
\label{lem:ofml-path-product-bound}
Under \eqref{eq:ofml-input-composition-lipschitz}, suppose that each module is Lipschitz continuous with constant $L_\tau(\theta)$.
Define $\Gamma_{p,v}(\theta)$ recursively for $v=1,\ldots,J_p$ by
\[
\Gamma_{p,v}(\theta)
:=
L_{\tau_{p,v}}(\theta)
\left(
L^{\mathrm{in}}_{p,v}
+
\sum_{w\in\pa_p(v)}\Gamma_{p,w}(\theta)
\right).
\]
Then
\[
\abs{f_\theta(x)-f_\theta(x')}
\le
\Gamma_{p,J_p}(\theta)\,d(x,x')
\qquad\forall x,x'\in\Xcal_p.
\]
Moreover,
\[
\Gamma_{p,J_p}(\theta)
=
\sum_{\substack{
v_0\to\cdots\to v_m=J_p\\
\text{directed paths in }p,\ m\ge0
}}
L^{\mathrm{in}}_{p,v_0}
\prod_{i=0}^{m}L_{\tau_{p,v_i}}(\theta).
\]
The sum includes paths starting at any node that has a directed path to $J_p$.
It also includes the path consisting only of $J_p$.
Each distinct node path is counted once, including paths with identical sequences of operation types.
A type appearing at several nodes of one path contributes one factor for each occurrence.
\end{slemma}

\begin{proof}
Define
\[
e_{p,v}
:=
\norm{
h_{\tau_{p,v}}(u_v^\theta(x);\theta_{\tau_{p,v}})
-
h_{\tau_{p,v}}(u_v^\theta(x');\theta_{\tau_{p,v}})
}.
\]
For a node without parents, \eqref{eq:ofml-input-composition-lipschitz}, the module's Lipschitz bound, and the definition of $\Gamma_{p,v}$ give
\[
e_{p,v}
\le
L_{\tau_{p,v}}(\theta)L^{\mathrm{in}}_{p,v}d(x,x')
=
\Gamma_{p,v}(\theta)d(x,x').
\]
Assume this bound holds for every parent of $v$.
The Euclidean norm of a concatenated vector is at most the sum of the norms of its blocks.
Apply the module's Lipschitz bound and \eqref{eq:ofml-input-composition-lipschitz}, followed by the parent bounds and the definition of $\Gamma_{p,v}$.
This gives
\[
\begin{aligned}
e_{p,v}
&\le
L_{\tau_{p,v}}(\theta)
\left(
L^{\mathrm{in}}_{p,v}d(x,x')
+
\sum_{w\in\pa_p(v)}e_{p,w}
\right)\\
&\le
\Gamma_{p,v}(\theta)d(x,x').
\end{aligned}
\]
Every parent has a smaller index than its child.
Induction over $v$ therefore proves this bound for every node.
The output of $J_p$ is $f_\theta$, which proves the stated Lipschitz bound for $f_\theta$.

Expand the recursion for $\Gamma_{p,J_p}$.
At each node, select either its external-input term or one of its parent terms.
Selecting an external-input term ends the expansion of that term.
The selected parent terms identify a directed path from the node where the expansion ends to $J_p$.
Thus paths start at intermediate nodes as well as nodes without parents.
The coefficient is $L^{\mathrm{in}}_{p,v_0}$, and the product contains the Lipschitz constant at every node on the path.
Selecting the external-input term at $J_p$ gives $L^{\mathrm{in}}_{p,J_p}L_{\tau_{p,J_p}}(\theta)$.
This term bounds the contribution of the terminal node's own external input.
Distinct node paths remain separate terms even when their operation types agree.
Repeated types along a path retain all their factors.
The DAG is finite, so this expansion gives the claimed finite sum.
\end{proof}

\begin{proof}[Proof of \Cref{thm:compact-ofml}]
By \Cref{ass:regularity,lem:ofml-path-product-bound}, $f_\theta$ is Lipschitz continuous on each $\Xcal_p$.
Points belonging to different protocols have distance $1$.
Thus $f_\theta|_{\Xtgt}\in C(\Xtgt)$.
By \Cref{ass:ofml-measurability}, the map $\theta\mapsto f_\theta|_{\Xtgt}$ is Borel measurable.
Together with \Cref{ass:correct-spec}, this gives a Borel measurable version of $f$ with values in $C(\Xtgt)$.

By \Cref{ass:correct-spec}, $\Pi_t$ is a version of $\Law(\boldsymbol\theta\mid\mathcal F_t)$.
\Cref{lem:ofml-conditional-pushforward} therefore identifies $\Pi_t^f$ as a version of $\Law(f\mid\mathcal F_t)$.
The integrability condition $\E\norm{f}_\infty^2<\infty$ follows from \Cref{ass:regularity}.
This verifies the posterior and integrability conditions in \Cref{main:assumption-5-1}.

For $x,x'\in\Xcal_p$, \Cref{lem:ofml-path-product-bound} gives $\abs{f(x)-f(x')}\le\Gamma_{p,J_p}(\boldsymbol\theta)d(x,x')$ almost surely.
For each target protocol, sum the coefficients $L^{\mathrm{in}}_{p,v_0}$ over the directed paths in \Cref{lem:ofml-path-product-bound}.
Let $Z_{\mathrm{path}}$ be the maximum of these sums over target protocols.
Each distinct node path contributes its coefficient once.
There are finitely many target protocols and finitely many paths in each protocol.
Thus $Z_{\mathrm{path}}$ is finite.
The path-product condition in \Cref{ass:regularity} yields $\max_{p\in\Ptgt}\Gamma_{p,J_p}(\boldsymbol\theta)\le Z_{\mathrm{path}}\Lambda$ almost surely.

If $x$ and $x'$ belong to different protocols, then $d(x,x')=1$.
The triangle inequality and this identity give
\[
\abs{f(x)-f(x')}
\le
\abs{f(x)}+\abs{f(x')}
\le
2\norm{f}_\infty d(x,x').
\]
Let $\omega$ be the identity function and set $L_f:=\max\!\left\{Z_{\mathrm{path}}\Lambda,\,2\norm{f}_\infty\right\}$.
The first term controls differences within each protocol.
The second term controls differences between points belonging to different protocols.
Consequently,
\[
\abs{f(x)-f(x')}
\le
L_f\,\omega(d(x,x'))
\qquad\forall x,x'\in\Xtgt
\quad\text{almost surely}.
\]
Moreover, $\E[L_f^2]\le Z_{\mathrm{path}}^2\E[\Lambda^2]+4\E\norm{f}_\infty^2<\infty$ by \Cref{ass:regularity}.
This verifies the modulus condition and completes the verification of \Cref{main:assumption-5-1}.

The joint measurability of $\DeltaIVR_t$ in \Cref{ass:ofml-measurability} gives \Cref{ass:measurability}.
The hypotheses also include \Cref{ass:noise,ass:cx-design}.
Applying \Cref{thm:general} proves both conclusions.
\end{proof}

\section{Uniform Consistency for Acquisition Functions Computed in Implementations}\label{sec:approx}

\Cref{thm:general,thm:compact-ofml} use \Cref{ass:cx-design}, which requires approximate maximization of the IVR gain.
An implementation maximizes an acquisition function approximating the exact IVR gain or the lower bound in \eqref{eq:lower-ivr}.
This appendix shows that the conclusions of these theorems hold for both choices.

\subsection{Acquisition Functions Approximating the IVR Gain}

\begin{slemma}\label{lem:approx}
Let $\widehat\Delta_t$ be an implemented acquisition function, and let $\widehat x_{t+1}\in\Xtgt$ be the selected point.
Write $\delta_t:=\norm{\widehat\Delta_t-\DeltaIVR_t}_\infty$ and $\eta_t^{\mathrm{opt}}:=\sup_{x\in\Xtgt}\widehat\Delta_t(x)-\widehat\Delta_t(\widehat x_{t+1})$.
Then
\[
\sup_{x\in\Xtgt}\DeltaIVR_t(x)-\DeltaIVR_t(\widehat x_{t+1})\le2\delta_t+\eta_t^{\mathrm{opt}}.
\]
Suppose that $\widehat x_{t+1}$, $\delta_t$, and $\eta_t^{\mathrm{opt}}$ are $\mathcal F_t$-measurable and that $\delta_t$ and $\eta_t^{\mathrm{opt}}$ converge to zero almost surely.
Then the selected points satisfy \eqref{eq:ivr-design} with the tolerance $\eta_t:=2\delta_t+\eta_t^{\mathrm{opt}}$.
\end{slemma}

\begin{proof}
For every $x\in\Xtgt$, $\DeltaIVR_t(x)\le\widehat\Delta_t(x)+\delta_t$ and $\DeltaIVR_t(\widehat x_{t+1})\ge\widehat\Delta_t(\widehat x_{t+1})-\delta_t$.
Taking the supremum over $x$ in the first inequality and subtracting the second inequality gives the bound.
The bound is the condition in \eqref{eq:ivr-design} with the stated tolerance.
\end{proof}

If $\delta_t$ or $\eta_t^{\mathrm{opt}}$ is not $\mathcal F_t$-measurable, any $\mathcal F_t$-measurable upper bound on $2\delta_t+\eta_t^{\mathrm{opt}}$ that converges to zero almost surely serves as the tolerance.
The upper semicontinuity of $\DeltaIVR_t$ in \Cref{ass:cx-design} and the other hypotheses of the theorems remain required.
\Cref{lem:approx} compares $\widehat\Delta_t$ with the IVR gain computed from the correct posterior of \Cref{main:assumption-5-1}.
Particle approximations used only to compute $\widehat\Delta_t$ therefore contribute to $\delta_t$.
If an approximation replaces the posterior used for prediction and updating, the hypotheses of the theorems must be verified for that approximation.

\subsection{Maximization of the IVR Lower Bound}

\begin{sproposition}\label{prop:lower-bound}
Define the lower bound
\[
\underline\Delta_t(x)
:=
\int_{\Xtgt}
\frac{k_t(x,x')^2}{s_t^2(x)+\xi^2}\,\nu(\mathrm dx').
\]
Replace \Cref{ass:cx-design} in \Cref{thm:general,thm:compact-ofml} by the following condition on point selection.
For every $t\ge0$, the point $x_{t+1}\in\Xtgt$ is $\mathcal F_t$-measurable and satisfies
\[
\underline\Delta_t(x_{t+1})
\ge
\sup_{x\in\Xtgt}\underline\Delta_t(x)-\eta_t^{\mathrm{lb}},
\]
where $\eta_t^{\mathrm{lb}}\ge0$ is $\mathcal F_t$-measurable and $\eta_t^{\mathrm{lb}}\to0$ almost surely.
Then the conclusions of \Cref{thm:general,thm:compact-ofml} hold without \Cref{ass:measurability} and without the joint-measurability requirement for $\DeltaIVR_t$ in \Cref{ass:ofml-measurability}.
\end{sproposition}

\begin{proof}
We first prove the conclusions of \Cref{thm:general}.
By \Cref{lem:posterior-moments}, $k_t$ and $s_t^2$ have versions that are continuous in the spatial arguments and measurable in the sample point.
Because $s_t^2(x)+\xi^2\ge\xi^2>0$ and $\nu$ is finite, $\underline\Delta_t$ is continuous in $x$ and jointly measurable.
Hence $\sup_{x\in\Xtgt}\underline\Delta_t(x)$ equals the supremum over a countable dense subset of $\Xtgt$ and is $\mathcal F_t$-measurable.

The first paragraph of the proof of \Cref{lem:ivr-drift} expresses $V_t-\E[V_{t+1}\mid\mathcal F_t]$ as an integral of conditional variances.
Because $x_{t+1}$ is $\mathcal F_t$-measurable, $\mathcal F_{t+1}=\sigma(\mathcal F_t,y_{t+1})$, and $y_{t+1}\mid f,\mathcal F_t\sim\mathcal N\bigl(f(x_{t+1}),\xi^2\bigr)$ by \Cref{ass:noise}.
The proof of \Cref{lem:noisy-ivr-bound}, with $y_{t+1}$ in place of $y^x$, therefore gives $V_t-\E[V_{t+1}\mid\mathcal F_t]\ge\underline\Delta_t(x_{t+1})$.
The telescoping and Tonelli arguments in the proof of \Cref{lem:ivr-drift} then give $\sum_{t=0}^{\infty}\underline\Delta_t(x_{t+1})<\infty$ almost surely.
With the condition on point selection,
\[
0\le\sup_{x\in\Xtgt}\underline\Delta_t(x)\le\underline\Delta_t(x_{t+1})+\eta_t^{\mathrm{lb}}\longrightarrow0
\quad\text{almost surely}.
\]

The proof of \Cref{lem:uniform-extinction} uses \Cref{lem:noisy-ivr-bound} and the convergence of the maximum IVR gain only to derive the contradiction in its last two sentences.
The left-hand side of \eqref{eq:uniform-extinction-integral} equals $\underline\Delta_{t_j}(\tilde x_j)$.
Thus the same proof with $\sup_x\underline\Delta_t(x)\to0$ in place of $\sup_x\DeltaIVR_t(x)\to0$ gives $\norm{s_t^2}_\infty\to0$ almost surely.
The second paragraph of the proof of \Cref{thm:general} uses only \Cref{main:assumption-5-1}, \Cref{lem:posterior-moments,lem:uniform-moments}, and $\norm{s_t^2}_\infty\to0$.
It therefore gives $\norm{\mu_t-f}_\infty\to0$ almost surely.

For OFML, \Cref{sec:ofml-proof} verifies \Cref{main:assumption-5-1} without \Cref{ass:cx-design} and without the joint measurability of $\DeltaIVR_t$.
The preceding argument therefore proves the conclusions of \Cref{thm:compact-ofml}.
\end{proof}

\section{Details of \texorpdfstring{\Cref{subsec:exp-transfer}}{Section 6.2}}
\label{sec:exp-gpmatched}

\Cref{sec:gpm-bench,sec:gpm-impl,sec:gpm-prespec} describe the benchmark, the implementations, and the settings fixed before the runs.
\Cref{sec:gpm-results} presents additional results.

\subsection{Benchmark Specification}
\label{sec:gpm-bench}

\paragraph{Protocols}
\Cref{tab:gpm-protocols} lists in text form the 17 protocols in \Cref{fig:exp-protocols}.

\begin{table}[t]
  \centering
  \caption{The 10 source protocols and the 7 target protocols in \Cref{subsec:exp-transfer}.
  An arrow from one type to another indicates that the output of the node of the first type is an input of the node of the second type.
  In $\mathsf M(\mathsf A,\mathsf B)$, the node of $\mathsf M$ receives the output of the node of $\mathsf A$ as its first input and that of $\mathsf B$ as its second input.
  The merge target protocols are the three rightmost protocols in \Cref{fig:exp-protocols}.
  The table lists them in their left-to-right order in that figure.}
  \label{tab:gpm-protocols}
  \small
  \begin{tabular}{@{}llc>{\raggedright\arraybackslash}p{0.62\textwidth}@{}}
    \toprule
    Role & Group & Depth & Protocols\\
    \midrule
    Source & Serial & 3 & $\mathsf A\to\mathsf{C}\to\mathsf{O}$, $\mathsf A\to\mathsf{D}\to\mathsf{O}$, $\mathsf B\to\mathsf{C}\to\mathsf{O}$, $\mathsf B\to\mathsf{D}\to\mathsf{O}$\\
    Source & Merge & 3 & $\mathsf M(\mathsf A,\mathsf B)\to\mathsf{O}$, $\mathsf M(\mathsf B,\mathsf A)\to\mathsf{O}$\\
    Source & Serial & 4 & $\mathsf A\to\mathsf{C}\to\mathsf{D}\to\mathsf{O}$, $\mathsf A\to\mathsf{D}\to\mathsf{C}\to\mathsf{O}$, $\mathsf B\to\mathsf{C}\to\mathsf{C}\to\mathsf{O}$, $\mathsf B\to\mathsf{D}\to\mathsf{D}\to\mathsf{O}$\\
    \addlinespace
    Target & Serial & 4 & $\mathsf A\to\mathsf{C}\to\mathsf{C}\to\mathsf{O}$, $\mathsf A\to\mathsf{D}\to\mathsf{D}\to\mathsf{O}$, $\mathsf B\to\mathsf{C}\to\mathsf{D}\to\mathsf{O}$, $\mathsf B\to\mathsf{D}\to\mathsf{C}\to\mathsf{O}$\\
    Target & Merge & 4 & $\mathsf M(\mathsf M(\mathsf A,\mathsf B),\mathsf A)\to\mathsf{O}$, $\mathsf M(\mathsf M(\mathsf B,\mathsf A),\mathsf B)\to\mathsf{O}$, $\mathsf M(\mathsf A,\mathsf M(\mathsf A,\mathsf B))\to\mathsf{O}$\\
    \bottomrule
  \end{tabular}
\end{table}

\paragraph{True functions and observations}
The true function of each type was drawn from the GP in \Cref{subsec:exp-metrics}. The signal standard deviation $0.70$ is the standard deviation of the function value at one input under the GP.
All nodes of the same type share one function, regardless of the protocol in which they appear.
Therefore, for each type, we drew the values of its function at the inputs received by all nodes of that type so that these values follow the joint distribution of the GP. We drew these values in order from parent nodes to child nodes. We drew each value conditioned on the values of the same type already drawn.
In each of the 10 source protocols, the source data $\mathcal D_0$ contain one observation at each of 8 grid points placed at nearly equal intervals among the 64 grid points. In total, the source data consist of 80 points.
The candidate points are the 448 points formed by the 64 grid points of each of the 7 target protocols.
Because the observations contain noise, observing a candidate point again still provides information. In this experiment, IVR was therefore allowed to select candidate points that had already been observed.
The realized observation noise is determined by the candidate point and by how many times that point has been observed before. It is the same for all surrogate models.

\paragraph{Input ranges in the target protocols}
A node that receives the external input $x$ is in the first layer, and the layer of every other node is one more than the largest layer of its parents.
The two nodes of $\mathsf M$ in each merge target protocol are in the second and third layers.
The second-layer $\mathsf M$ receives the same inputs as the $\mathsf M$ of the merge source protocols of depth 3.
On the 12 true functions used for the results in \Cref{subsec:exp-transfer}, we measured how far the inputs of target nodes lie from the inputs of source nodes.
We examined the third-layer $\mathsf C$ and $\mathsf D$ of the serial target protocols, the third-layer $\mathsf M$ of the merge target protocols, and $\mathsf O$ of all target protocols.
The external input $x$ takes the values of the 64 grid points.
The output of each node is an input of its child nodes, so the input that a node receives consists of the outputs of its parent nodes and is determined by $x$.
The reference points of a type are the inputs that all its nodes in the source protocols receive.
For each input that an examined node receives, we computed the Euclidean distance to the nearest reference point of the same type.
We compared this distance with $0.5$, the length-scale shared by $\mathsf C$, $\mathsf D$, $\mathsf M$, and $\mathsf O$.
In the serial target protocols, this distance was at most $0.5$ for every input of the third-layer $\mathsf C$ and $\mathsf D$ and of $\mathsf O$.
For the third-layer $\mathsf M$, the fraction of inputs with this distance above $0.5$ was $0.27$ on average over the 12 functions.
In the merge target protocols, inputs of $\mathsf O$ with this distance above $0.5$ occurred for only one true function, and there were only three such inputs.

\paragraph{Random seeds}
The 34 true functions drawn for \Cref{subsec:exp-transfer} and the 4 true functions for the hyperparameter search were drawn from disjoint sets of random seeds.

\subsection{Implementation of the Surrogate Models}
\label{sec:gpm-impl}

This subsection describes the inference methods for the two GP surrogate models and the training methods and architectures of the four deep-ensemble surrogate models.
\Cref{alg:shared} shows the computation shared by the surrogate models. \Cref{alg:gpn,alg:ens} give the procedures of OFML-GPN and of the deep-ensemble surrogate models.

\begin{algorithm}[t]
\caption{Computation shared by the surrogate models in \Cref{subsec:exp-transfer}}\label{alg:shared}
\begin{algorithmic}[1]
\Require sequence $\mathcal D_0$ of the 80 source observations; candidate set $\Xtgt$; $T=200$
\Ensure posterior mean $\mu_t\colon\Xtgt\to\mathbb R$ and covariance $k_t\colon\Xtgt^2\to\mathbb R$ for $t=0,\ldots,T$
\State \Call{Init}{\,}
\State $\mathcal D\gets\mathcal D_0$ \Comment{sequence of pairs $(x,y)$; repeated points are kept}
\State \Call{Absorb}{$\mathcal D_0$}
\For{$t=0,\ldots,T$}
  \State $(\mu_t,k_t)\gets$ \Call{Predict}{\,}
  \If{$t=T$} \textbf{break} \EndIf
  \State $x_{t+1}\gets\operatorname*{arg\,max}_{x\in\Xtgt}\underline\Delta_t(x)$ \Comment{\eqref{eq:lower-ivr}; observed points remain candidates}
  \State $y_{t+1}\gets$ noisy observation of the true function at $x_{t+1}$
  \State append $(x_{t+1},y_{t+1})$ to $\mathcal D$
  \State \Call{Absorb}{$((x_{t+1},y_{t+1}))$}
\EndFor
\end{algorithmic}
\end{algorithm}

\subsubsection{GP Surrogate Models}
\label{sec:gpm-gp}

\Cref{alg:gpn} gives the procedures of OFML-GPN.

\begin{algorithm}[t]
\caption{Procedures of OFML-GPN ($M=16{,}384$). \Cref{alg:shared} calls these procedures. $L_n(v)$ is the marginal likelihood of the first $n$ observations in $\mathcal D$ under the GP regression of $\mathsf O$ with inputs $v$. $v^{(m)}$ denotes the node outputs given by $u^{(m)}$.}\label{alg:gpn}
\begin{algorithmic}[1]
\Procedure{Init}{\,}
  \State draw $u^{(m)}\sim\mathcal N(0,I)$ and set $\ell^{(m)}\gets0$ for $m=1,\ldots,M$ \Comment{whitened coordinates}
  \State $\beta_\tau\gets0.30$ for each type $\tau\in\{\mathsf A,\mathsf B,\mathsf C,\mathsf D,\mathsf M\}$; $n\gets0$
\EndProcedure
\Procedure{Absorb}{$\mathcal E$}
  \For{each $(x,y)$ in $\mathcal E$, in order}
    \State $n\gets n+1$; $\ell^{(m)}\gets\ell^{(m)}+\log L_n(v^{(m)})-\log L_{n-1}(v^{(m)})$ for all $m$
    \State $w^{(m)}\gets\exp(\ell^{(m)})/\sum_{m'}\exp(\ell^{(m')})$ for all $m$
    \If{$1/\sum_m(w^{(m)})^2<0.65M$}
      \State resample $\{u^{(m)}\}_{m=1}^M$ systematically by $\{w^{(m)}\}_{m=1}^M$; $\ell^{(m)}\gets0$ and $w^{(m)}\gets1/M$ for all $m$
      \For{12 repetitions}
        \For{8 sweeps}
          \State update the block of each type $\tau$ of every $u^{(m)}$ by pCN with step size $\beta_\tau$
          \State redraw the values of the parents of $\mathsf O$ at unobserved grid points
        \EndFor
        \State $\beta_\tau\gets\min\{0.5,\max\{0.002,\beta_\tau\exp(0.5(a_\tau-0.23))\}\}$ for each $\tau$ \Comment{$a_\tau$: acceptance rate}
      \EndFor
    \EndIf
  \EndFor
\EndProcedure
\Procedure{Predict}{\,}
  \State $\mu^{(m)},k^{(m)}\gets$ posterior mean and covariance of the GP regression of $\mathsf O$ given $v^{(m)}$ and $\mathcal D$
  \State $\mu(x)\gets\sum_m w^{(m)}\mu^{(m)}(x)$
  \State $k(x,x')\gets\sum_m w^{(m)}[k^{(m)}(x,x')+\{\mu^{(m)}(x)-\mu(x)\}\{\mu^{(m)}(x')-\mu(x')\}]$
  \State \Return $(\mu,k)$
\EndProcedure
\end{algorithmic}
\end{algorithm}

OFML-GPN treats the outputs of nodes whose type is not $\mathsf O$ as latent variables.
It approximates their posterior distribution by sequential Monte Carlo.
In the experiments, these variables comprised each such node's outputs at the 64 grid points of the external input $x$.
Whitened coordinates $u$ represent these latent variables with joint prior distribution $\mathcal N(0,I)$.
The latent node outputs $v$ are a deterministic function of $u$.
Given $v$, the inputs of $\mathsf O$ are fixed, and GP regression marginalizes the function of $\mathsf O$ analytically.
This marginal likelihood, evaluated at the node outputs of a particle, is the likelihood of that particle.

Each pCN update changes the block $u_\tau$ of type $\tau$ of one particle and keeps the other blocks fixed \citep{Cotter2013}.
With the other blocks fixed, write $L_n(u_\tau)$ for $L_n$ as a function of $u_\tau$.
In whitened coordinates, the prior density of $u_\tau$ is the standard normal density $\phi$, and the target density is $\pi_n(u_\tau)\propto L_n(u_\tau)\,\phi(u_\tau)$.
The proposal distribution is $q_\tau(u_\tau'\mid u_\tau)=\mathcal N\bigl(u_\tau';\sqrt{1-\beta_\tau^2}\,u_\tau,\beta_\tau^2 I\bigr)$, and $\beta_\tau\in(0,1]$ sets the size of the move.
The Metropolis--Hastings acceptance probability is $\min\{1,\pi_n(u_\tau')\,q_\tau(u_\tau\mid u_\tau')/[\pi_n(u_\tau)\,q_\tau(u_\tau'\mid u_\tau)]\}$.
This proposal distribution satisfies $\phi(u_\tau)\,q_\tau(u_\tau'\mid u_\tau)=\phi(u_\tau')\,q_\tau(u_\tau\mid u_\tau')$, so the prior and proposal terms cancel.
The acceptance probability therefore reduces to $\min\{1,L_n(u_\tau')/L_n(u_\tau)\}$.

Line 13 of \Cref{alg:gpn} redraws the values of the parent node of $\mathsf O$ in each protocol at the grid points not observed in that protocol.
The values are drawn from the exact conditional prior given the values of the same node at the observed points and all other latent variables.
Conditional on these variables, the redrawn values do not affect the likelihood of the current observations.
The redraw therefore samples their conditional posterior directly.
The predictive distribution of the true terminal response $f_p$ is the mixture of the GP-regression posteriors of the particles, weighted by the particle weights.

Independent-GPN assigns separate functions of every type, including $\mathsf O$, to each target protocol.
Nodes of the same type share one function within a protocol.
Each type has the same GP prior as in OFML-GPN.
In the experiments, Independent-GPN kept a separate particle set for each target protocol and updated each set by the inference method of OFML-GPN.
Each observation was assimilated only into the particle set for its protocol.
Each set used 65,536 particles and 384 pCN sweeps after resampling.
Independent-GPN did not use $\mathcal D_0$, so each particle set started from the prior.

\subsubsection{Deep-Ensemble Surrogate Models}
\label{sec:gpm-nn}

\Cref{alg:ens} gives the procedures of the deep-ensemble surrogate models.

\begin{algorithm}[t]
\caption{Procedures of the deep-ensemble surrogate models ($M=5$). \Cref{alg:shared} calls these procedures.}\label{alg:ens}
\begin{algorithmic}[1]
\Procedure{Init}{\,}
  \State draw initial parameters $\theta_0^{(1)},\ldots,\theta_0^{(M)}$
  \State $(\bar y,\sigma_y)\gets$ mean and standard deviation of the 80 responses in $\mathcal D_0$ \Comment{fixed for all $t$}
\EndProcedure
\Procedure{Absorb}{$\mathcal E$}
  \For{$m=1,\ldots,M$}
    \State $\theta^{(m)}\gets\theta_0^{(m)}$ and reinitialize Adam \Comment{retrain on all of $\mathcal D$}
    \State apply 400 Adam updates to $|\mathcal D|^{-1}\sum_{(x,y)\in\mathcal D}\{f_{\theta^{(m)}}(x)-(y-\bar y)/\sigma_y\}^2$ plus the $L_2$ penalty
  \EndFor
\EndProcedure
\Procedure{Predict}{\,}
  \State $g^{(m)}(x)\gets\bar y+\sigma_y f_{\theta^{(m)}}(x)$ for $m=1,\ldots,M$
  \State $\mu(x)\gets M^{-1}\sum_m g^{(m)}(x)$
  \State $k(x,x')\gets(M-1)^{-1}\sum_m\{g^{(m)}(x)-\mu(x)\}\{g^{(m)}(x')-\mu(x')\}+10^{-8}\delta_{xx'}$
  \State \Return $(\mu,k)$
\EndProcedure
\end{algorithmic}
\end{algorithm}

The members of each surrogate model differ only in their initial weights.
Retraining every member from its initial values at each $t$ followed \citet{Ash2020}.
Adam added the gradient of the $L_2$ penalty to the loss gradient of every trainable parameter, including the biases.
The standard deviation $\sigma_y$ was computed with denominator 80.

The networks $f_\theta$ in \Cref{alg:ens} have the architectures in \Cref{tab:gpm-nn}.
The hidden layers of all multilayer perceptrons use the ReLU activation.
Within each of OFML-Ens and Shared-DAG-Ens, all multilayer perceptrons have the same number of hidden layers and the same width.
\Cref{tab:gpm-hpo} gives these values.
In OFML-Ens and Shared-DAG-Ens, $r$ denotes the dimension of the latent states.
Each member of these two models has one affine map from $x$ to an $r$-dimensional vector, called the latent state of $x$.
In every protocol, all nodes of type $\mathsf A$ or $\mathsf B$ share this map.
The type embeddings of Shared-DAG-Ens are four-dimensional parameters learned for each type.

DAGNN-Cat-Ens is DAGNN \citep{Thost2021} with the attention-based aggregation of the parent states replaced by concatenation.
One input node that receives $x$ is added to each protocol.
The input node is a parent of every node of type $\mathsf A$ or $\mathsf B$.
The initial state of the input node is an affine image of $x$.
The initial state of every other node, including the node of $\mathsf O$, is a vector learned for its type.
Each DAGNN layer has its own affine map and its own GRU.
Each DAGNN layer updates the state of every node once, in a topological order of the DAG.
When a node is updated, the GRU input is the state of that node after the previous DAGNN layer.
In the first DAGNN layer, the GRU input is the initial state of the node.
For a node with parents, the parent states already updated in the current DAGNN layer are concatenated.
The affine map of the current DAGNN layer transforms the concatenated vector.
The transformed vector is the hidden-state input of the GRU.
The input node has no parents, so its GRU hidden state is the zero vector.
The GRU output is the new state of the node.
The readout is a multilayer perceptron with one hidden layer.

For a given $x$, each Shared-Ens member gives the same prediction to protocols with the same type counts, such as $\mathsf M(\mathsf A,\mathsf B)\to\mathsf O$ and $\mathsf M(\mathsf B,\mathsf A)\to\mathsf O$.
Shared-Ens omits the count of $\mathsf O$ from its input because every protocol contains exactly one node of $\mathsf O$.

\begin{table}[t]
  \centering
  \caption{Architectures of one member of each deep-ensemble surrogate model in \Cref{subsec:exp-transfer}.
  The table lists the networks that compute the node states and the network that outputs the response.
  In OFML-Ens and Shared-DAG-Ens, the multilayer perceptron of $\mathsf O$ has the same number of hidden layers and width as the other multilayer perceptrons of the model.}
  \label{tab:gpm-nn}
  \small
  \setlength{\tabcolsep}{3pt}
  \begin{tabular}{@{}>{\raggedright\arraybackslash}p{\dimexpr0.18\textwidth-1.5\tabcolsep\relax}>{\raggedright\arraybackslash}p{\dimexpr0.23\textwidth-1.5\tabcolsep\relax}>{\raggedright\arraybackslash}p{\dimexpr0.34\textwidth-1.5\tabcolsep\relax}>{\raggedright\arraybackslash}p{\dimexpr0.25\textwidth-1.5\tabcolsep\relax}@{}}
    \toprule
    Surrogate model & Networks for the node states & Input of these networks & Output of the response\\
    \midrule
    OFML-Ens & A separate multilayer perceptron for each of $\mathsf A$, $\mathsf B$, $\mathsf C$, $\mathsf D$, $\mathsf M$ & For the first-layer types, the latent state of $x$; for $\mathsf C$ and $\mathsf D$, the latent state of the parent; for $\mathsf M$, the concatenation of the latent states of the first and second parents & The multilayer perceptron of $\mathsf O$ maps the latent state of the parent to the response\\
    \addlinespace
    Shared-DAG-Ens & One multilayer perceptron shared by $\mathsf A$, $\mathsf B$, $\mathsf C$, $\mathsf D$, $\mathsf M$ & The concatenation of two latent states and the type embedding. For the first-layer types, the two latent states are the latent state of $x$ and the zero vector. For $\mathsf{C}$ and $\mathsf{D}$, they are the latent state of the parent and the zero vector. For $\mathsf M$, they are the latent states of the first and second parents & A separate multilayer perceptron of $\mathsf O$ maps the latent state of the parent to the response\\
    \addlinespace
    DAGNN-Cat-Ens & A GRU and an affine map for each DAGNN layer, shared by all types including $\mathsf O$ & The concatenation of the latent states of the first and second parents. For a node with one parent, the second vector in the concatenation is the zero vector & A readout maps the initial state and the states after each DAGNN layer of the node of $\mathsf O$ to the response\\
    \addlinespace
    Shared-Ens & None; one multilayer perceptron predicts from the protocol-level input & The 6-dimensional vector of $x$ and the numbers of uses of $\mathsf A$, $\mathsf B$, $\mathsf C$, $\mathsf D$, $\mathsf M$ & The same multilayer perceptron outputs the response\\
    \bottomrule
  \end{tabular}
\end{table}

\subsection{Pre-Specified Settings}
\label{sec:gpm-prespec}

We fixed the hyperparameters and the exclusion criterion below before the runs.

\paragraph{Hyperparameters}
\label{sec:gpm-hpo}
The two GP surrogate models used the hyperparameters of the GP from which the true functions were drawn (\Cref{subsec:exp-metrics}).
The kernel is the squared exponential kernel, and the signal standard deviation is $0.70$.
The length-scale is $1.0$ for the first-layer types and $0.5$ for the other types, and the noise variance is $\gamma^2=\xi^2=0.01$.
For each deep-ensemble surrogate model, we selected a hyperparameter configuration $\psi$ by Bayesian optimization, which used GP regression and expected improvement \citep{Snoek2012}.
A configuration $\psi$ consists of the values of the number of hidden layers, the width, the latent dimension, the learning rate, and the $L_2$ regularization coefficient in \Cref{tab:gpm-hpo}.
For DAGNN-Cat-Ens and Shared-Ens, $\psi$ consists of the four values other than the latent dimension.
\Cref{sec:gpm-nn} describes these quantities.
In each trial, we fixed one configuration $\psi$, ran \Cref{alg:shared} up to $T=200$ on each of the 4 search functions, and computed the objective
\[
\frac{1}{20}
\sum_{j=1}^{4}
\sum_{t\in\{0,20,50,100,200\}}
\NRMSE_j(t;\psi).
\]
Here $\NRMSE_j(t;\psi)$ denotes $\NRMSE(t)$ defined in \Cref{subsec:exp-metrics}, evaluated on search function $j$ with configuration $\psi$.
Each NRMSE value used all 448 candidate points of the target protocols.
The 4 search functions were disjoint from the 34 true functions drawn for \Cref{subsec:exp-transfer} and were shared by the four deep-ensemble surrogate models.
The search consisted of 80 trials.
Trials 1 to 10 drew $\psi$ from a Sobol sequence, and trials 11 to 80 selected $\psi$ by Bayesian optimization.
Trials 1 to 50 restricted the number of hidden layers to 1--3, the width to 8--64, the learning rate to $[10^{-3},3\times10^{-2}]$, and the $L_2$ regularization coefficient to $[10^{-6},10^{-3}]$.
Trials 51 to 80 used the search ranges in \Cref{tab:gpm-hpo}.
The runs in \Cref{subsec:exp-transfer} used the configuration of the trial whose objective value was the smallest among the 80 trials.
The number of Adam updates was fixed to 400.

\begin{table}[t]
  \centering
  \caption{Search ranges and selected values of the hyperparameters of the deep-ensemble surrogate models in \Cref{subsec:exp-transfer}.
  Log indicates a logarithmic scale. The number of hidden layers, the width, and the latent dimension were rounded to integers.
  Each deep-ensemble surrogate model used 400 Adam updates, 5 members, and the whole $\mathcal D_t$ as one batch.
  For DAGNN-Cat-Ens, the number of hidden layers specifies the number of DAGNN layers.
  The width specifies both the dimension of the GRU state and the width of the readout.
  A dash indicates that the latent dimension was not searched for that surrogate model.}
  \label{tab:gpm-hpo}
  \footnotesize
  \setlength{\tabcolsep}{2.5pt}
  \begin{tabular}{@{}ll c cccc@{}}
    \toprule
     & & & \multicolumn{4}{c}{Selected value}\\
    \cmidrule(l){4-7}
    Quantity & Scale & Search range & OFML-Ens & Shared-DAG-Ens & DAGNN-Cat-Ens & Shared-Ens\\
    \midrule
    Number of hidden layers & Linear & 1--5 & 2 & 2 & 3 & 3\\
    Width & Log & 4--128 & 63 & 126 & 14 & 56\\
    Latent dimension & Linear & 2--8 & 3 & 2 & -- & --\\
    Learning rate & Log & $[10^{-3},10^{-1}]$ & $6.2\times10^{-3}$ & $9.2\times10^{-3}$ & $1.0\times10^{-2}$ & $3.0\times10^{-2}$\\
    $L_2$ regularization coefficient & Log & $[10^{-6},10^{-2}]$ & $6.8\times10^{-6}$ & $2.5\times10^{-6}$ & $1.0\times10^{-6}$ & $1.0\times10^{-3}$\\
    \bottomrule
  \end{tabular}
\end{table}

\paragraph{Exclusion and replacement of true functions}
\label{sec:gpm-exclusion}
\label{sec:gpm-replacement}
We fixed the following exclusion criterion before the runs.
At assimilation $n$, let $w^{(m)}$ denote the normalized particle weights after the likelihood update and before resampling.
Define
\[
\zeta_n
=
\frac{1}{M}
\frac{1}{\sum_{m=1}^{M}(w^{(m)})^2},
\]
where $M$ is the number of particles.
The second factor is the effective sample size compared with $0.65M$ in line 8 of \Cref{alg:gpn}.
For Independent-GPN, both the weights and $M$ refer to the particle set that assimilates the observation.
We recorded $\zeta_n$ at every assimilation under both GP surrogate models. For OFML-GPN, these assimilations included those of $\mathcal D_0$.
A true function meets the exclusion criterion if $\min_n\zeta_n<10^{-3}$ under OFML-GPN or Independent-GPN.
The exclusion criterion uses only the effective sample size and uses neither prediction errors nor true responses.
A function that meets the exclusion criterion is excluded from the comparison of all surrogate models.
We first drew 12 true functions.
Following the plan fixed before the runs, we replaced the 4 functions that met the exclusion criterion under OFML-GPN with 4 newly drawn functions.
All surrogate models were run on these 16 functions.
After seeing these results and before any run on further functions, we fixed the procedure for further replacement.
We kept the 7 functions among these 16 that met the exclusion criterion under neither GP surrogate model.
We drew further functions in order.
In drawing order, we added the first 5 further functions that met the exclusion criterion under neither GP surrogate model.
Only the drawing order and the exclusion criterion determined the selection of the 12 functions used for the results.
\Cref{tab:gpm-exclusion} gives the numbers of functions.
Independent-GPN was run on all first 16 functions and on the 8 further functions that did not meet the exclusion criterion under OFML-GPN.
Under Independent-GPN, 1 of the first 16 functions met the exclusion criterion.
This function also met the exclusion criterion under OFML-GPN.
None of the 8 further functions met the exclusion criterion under Independent-GPN.
The run logs in \Cref{sec:repro} record the decision for each function.
\Cref{sec:gpm-diagnostics} analyzes the excluded functions.

\begin{table}[t]
  \centering
  \caption{Numbers of true functions that met the exclusion criterion in \Cref{subsec:exp-transfer}.
  The two OFML-GPN columns distinguish the assimilation stages at which the exclusion criterion was first met.
  Up to 8 OFML-GPN runs were executed in parallel, so OFML-GPN was run on 18 further functions.
  Of the 8 further functions that did not meet the exclusion criterion under OFML-GPN, 3 were drawn after the fifth added function.
  These 3 functions were not used for the results.
  In each row, the number used for the results equals the number drawn minus the numbers in the two OFML-GPN columns and in the column of functions drawn after the fifth added function.}
  \label{tab:gpm-exclusion}
  \small
  \begin{tabular}{@{}l >{\centering\arraybackslash}p{0.08\textwidth} >{\centering\arraybackslash}p{0.12\textwidth} >{\centering\arraybackslash}p{0.17\textwidth} >{\centering\arraybackslash}p{0.17\textwidth} >{\centering\arraybackslash}p{0.10\textwidth}@{}}
    \toprule
     & & \multicolumn{2}{>{\centering\arraybackslash}p{0.31\textwidth}}{Met the exclusion criterion under OFML-GPN} & & \\
    \cmidrule(lr){3-4}
    True functions & Drawn & Assimilation of $\mathcal D_0$ & Assimilation of target-protocol observations & Drawn after the fifth added function and not used & Used for the results\\
    \midrule
    First 16 functions & 16 & 4 & 5 & 0 & 7\\
    Further functions & 18 & 5 & 5 & 3 & 5\\
    \addlinespace
    Total & 34 & 9 & 10 & 3 & 12\\
    \bottomrule
  \end{tabular}
\end{table}

\subsection{Additional Results}
\label{sec:gpm-results}

\paragraph{Differences between surrogate models}
\label{sec:gpm-paired}
\Cref{tab:gpm-paired} reports the mean and 95\% interval of the per-function difference for five pairs of surrogate models on the 12 true functions, for the All, Serial, and Merge groups.
Each of the five pairs in the table differs in one or two of the elements in \Cref{tab:exp-factors}, and the pairs separate the contributions of the components of OFML.

We first consider the contribution of parameter sharing.
For OFML-GPN and Independent-GPN, the 95\% intervals of the differences in $\NRMSE(0)$ and $\AULC(T)$ excluded zero in every group, and those in $\NRMSE(T)$ included zero in every group.
Independent-GPN does not share parameters, so it cannot use the source observations to predict the target protocols.
Therefore, the contribution of sharing lies in transferring the source observations to the target protocols.
This contribution appears while the target observations are few, and after 200 observations no difference from a surrogate model that uses only the target observations was detected.

We next consider the contribution of composition.
For Shared-DAG-Ens and Shared-Ens, the differences were negative for all three metrics in all three groups, with 95\% intervals excluding zero.
That is, unlike the contribution of sharing, the contribution of composition remained in the final error.
The reason is that Shared-Ens, which does not compose modules, cannot distinguish protocols with the same type counts, such as $\mathsf B\to\mathsf C\to\mathsf D\to\mathsf O$ and $\mathsf B\to\mathsf D\to\mathsf C\to\mathsf O$.
Therefore, composition is necessary to represent the target protocols separately, and additional observations cannot replace it.

We then consider the contribution of the function model and the inference for each type.
For OFML-GPN and OFML-Ens, the differences were also negative for all three metrics in all three groups, with 95\% intervals excluding zero.
The two surrogate models have the same composition and sharing, so this difference comes from the function model and the inference.
That is, the part of the error of OFML-Ens that exceeds the error of OFML-GPN comes from the approximation by multilayer perceptrons and deep ensembles, not from the structure of OFML.

Finally, we consider the contribution of type-specific modules.
For OFML-Ens and Shared-DAG-Ens, the 95\% intervals of the differences in $\NRMSE(0)$ and $\NRMSE(T)$ included zero in every group, and the difference in $\AULC(T)$ was positive in All and Merge.
Therefore, in this benchmark, separate modules for each type did not reduce the error, and they slowed adaptation in the merge group.
For OFML-Ens and DAGNN-Cat-Ens, $\NRMSE(0)$ differed in All and Serial.
However, DAGNN-Cat-Ens also differs from OFML-Ens in its function model, so this difference does not show a contribution of type-specific modules.

In summary, the advantage of OFML in this benchmark came from two components.
Sharing reduced the error while observations were few.
Composition remained effective up to the final error.
No contribution of type-specific modules was observed.

\begin{table}[t]
  \centering
  \caption{Per-function differences for five pairs of the surrogate models in \Cref{subsec:exp-transfer}.
  The differences were computed on the 12 true functions.
  Here $T=200$.
  Each difference is the value of the surrogate model written first minus the value of the surrogate model written second.
  The 95\% interval is the percentile bootstrap interval of the mean difference over true functions.
  Count is the number of the 12 true functions on which the surrogate model written first has the smaller value.}
  \label{tab:gpm-paired}
  \footnotesize
  \setlength{\tabcolsep}{2pt}
  \input{generated_exp/table_obj1_paired_en.tex}
\end{table}

\paragraph{Diagnostics of the OFML-GPN particle approximation}
\label{sec:gpm-diagnostics}
Because OFML-GPN is a correctly specified surrogate model, the expected 95\% coverage averaged over true functions is $0.95$ when its posterior distribution is computed exactly.
This paragraph examines the particle approximation of OFML-GPN with $\zeta_n$ and the 95\% coverage.

Of the first 16 true functions that we ran, 4 met the exclusion criterion during the assimilation of $\mathcal D_0$, before any target protocol was observed.
On 2 of them, we examined whether increasing the number of particles or the number of pCN sweeps would avoid the exclusion criterion.
One setting increased only the number of particles from 16,384 to 65,536, and the other increased only the number of pCN sweeps after each resampling from 96 to 384.
In both settings, we computed only the initialization, the assimilation of $\mathcal D_0$, and the prediction at $t=0$.
In both settings, both functions met the exclusion criterion, and the 95\% coverage at $t=0$ ranged from $0.36$ to $0.60$, below the nominal $0.95$.
Thus, on these 2 functions, four times as many particles or sweeps did not remove the degeneracy of the particle approximation.

On the 12 true functions used for the results, the mean coverage in All was $0.93$ at $t=0$ and decreased to $0.81$ at $t=200$.
Thus, the error of the particle approximation grew as observations accumulated.
In Serial, OFML-GPN had a higher mean final error than Independent-GPN (\Cref{tab:exp-transfer}).
This difference is probably due to the error of the particle approximation.
After obtaining the results, we identified the 3 functions with the lowest coverage in Serial at $t=200$ and kept them among the 12 functions used for the results.
Their coverage values were $0.40$, $0.64$, and $0.65$, and the final errors of OFML-GPN in Serial were $0.97$, $0.37$, and $0.72$.
Over the other 9 functions, the mean final error in Serial was $0.18$ for OFML-GPN and $0.21$ for Independent-GPN, so OFML-GPN had the smaller error.
Thus, the higher mean final error of OFML-GPN in Serial came from the errors on these 3 low-coverage functions.

\section{Details of \texorpdfstring{\Cref{subsec:exp-acquisition}}{Section 6.3}}
\label{sec:acq-details}

\Cref{sec:acq-bench,sec:acq-impl,sec:acq-prespec} describe the benchmark, the implementations, and the hyperparameter selection and exclusion criterion.
\Cref{sec:acq-results} presents additional results.
A run is the computation for one surrogate model, one true function, and one acquisition function.
Each run starts from the 8 initial observations in $\mathcal D_0$ and adds $T=92$ observations.
The total number of observations at $t$ is $t+8$, and the totals 8, 50, and 100 in the tables correspond to $t=0$, $42$, and $92$, respectively.

\subsection{Benchmark Specification}
\label{sec:acq-bench}

The types, protocols, true functions, and candidate points are given in \Cref{subsec:exp-acquisition}.
Each of the 8 protocols whose second-layer type is $\mathsf{G1}$, $\mathsf{G3}$, $\mathsf{G5}$, or $\mathsf{G7}$ has one initial point.

\subsection{Implementation of the Surrogate Models and Acquisition Functions}
\label{sec:acq-impl}

\paragraph{Surrogate models}
OFML-GPN is the OFML-GPN of \Cref{sec:gpm-gp} with only the terminal type and the updated blocks changed.
The terminal type is $\mathsf Q$, and each sweep updates the 10 blocks of $\mathsf E$, $\mathsf F$, and $\mathsf{G1},\ldots,\mathsf{G8}$ in turn.
The number of particles, the resampling threshold, and the number of pCN sweeps after resampling are the same as in \Cref{sec:gpm-gp}.

OFML-Ens applies the architecture and the training method of the OFML-Ens of \Cref{sec:gpm-nn} to the types $\mathsf E$, $\mathsf F$, $\mathsf{G1},\ldots,\mathsf{G8}$, and $\mathsf Q$.
Its hyperparameters are the values selected in \Cref{sec:acq-ens-selection}.
At each $t$, responses were standardized by the mean and standard deviation of all responses observed up to $t$.

For OFML-Ens, $k_t$ and $s_t^2$ in IVR are defined as in \Cref{sec:gpm-nn}.
Let $\tilde s_t^2(x)$ be the sum of the squared deviations of the five member predictions from their mean, divided by 5.
Then $s_t^2(x)=\tfrac54\tilde s_t^2(x)+10^{-8}$.
As stated in the caption of \Cref{fig:exp-acquisition}, MaxVar, the 95\% coverage, and the posterior variance in \Cref{fig:exp-acquisition}(b) use $\tilde s_t^2$.
With $\tilde s_t(x)$ the square root of $\tilde s_t^2(x)$, the 95\% coverage of OFML-Ens uses the interval $\mu_t(x)\pm1.96\,\tilde s_t(x)$.

\paragraph{Acquisition functions}
Space-filling visits the 16 protocols in a fixed cyclic order and selects one grid point per visit.
The selected point is the first unobserved grid point of that protocol in the base-2 van der Corput order.

\subsection{Hyperparameter Selection and Exclusion Criterion}
\label{sec:acq-prespec}

\paragraph{Selection of the OFML-Ens hyperparameters}
\label{sec:acq-ens-selection}
Let $\psi$ denote a hyperparameter configuration of OFML-Ens.
It consists of the values of the 7 quantities in \Cref{tab:acq-ens-selection}, and the number of members was fixed to 5.
The selection of $\psi$ used 32 true functions drawn separately from the 16 used for the results in \Cref{fig:exp-acquisition}, split into 16 search functions and 16 confirmation functions.
In the search range, the configuration with width 1024 and one hidden layer was added after preliminary runs on the 16 functions used for the results.
For $\psi$ and a set of 16 true functions, let $R(\psi)$ be the mean of $\AULC(T)$ over the four acquisition functions and the 16 functions.
We searched for $\psi$ minimizing $R$ on the search functions with 32 trials drawn from a Sobol sequence and 64 trials selected by Bayesian optimization with GP regression and expected improvement.
The number of weights of one member was counted as $11\{(\text{number of hidden layers}-1)\times\text{width}^2+2\times\text{latent dimension}\times\text{width}\}$, and only $\psi$ with at most $5.2\times10^{6}$ weights were searched.
Among the trials, we used only those whose root mean squared error of the posterior mean against the observed responses at the observed points was at most 1.25 times the noise standard deviation.
This error was averaged over 50 to 100 observations and over the 64 runs of the trial.
From these trials, we took 5 configurations in increasing order of the GP posterior mean at the end of the search, skipping any configuration close to one already taken.
From these trials, we also took the configuration with the smallest $R$ among the first 32 trials.
The 8 candidates consisted of these 6 configurations and the 1-layer and 4-layer configurations fixed before the search.
Two configurations are close if they have the same width, number of hidden layers, and weight initialization, and the Euclidean distance between the remaining quantities, each mapped from its search range to $[0,1]$ on the scale in \Cref{tab:acq-ens-selection}, is at most $0.1$.
We recomputed $R$ on the confirmation functions for these candidates and selected the candidate with the smallest value.
We also selected $\psi$ by the same search with the $\AULC(T)$ of Random alone as the objective, and this search was planned before the trials started.

\begin{table}[t]
  \centering
  \caption{Search ranges and hyperparameter values of OFML-Ens in \Cref{subsec:exp-acquisition}.
  The width was chosen from powers of 2. The number of members was fixed to 5.
  The table shows the selected configuration, the candidate with the second smallest confirmation $R$, and the 1-layer and 4-layer configurations fixed before the search. The confirmation $R$ of the remaining 4 candidates ranged from $0.860$ to $0.878$.
  The selected configuration is the one selected by the mean $\AULC(T)$ over the four acquisition functions.
  A dash indicates a value that was not computed.
  Bold marks the confirmation $R$ of the selected configuration.}
  \label{tab:acq-ens-selection}
  \footnotesize
  \setlength{\tabcolsep}{2.5pt}
  \begin{tabular}{@{}ll l cccccc@{}}
    \toprule
    Quantity & Scale & Search range & \shortstack{Selected\\configuration} & \shortstack{Second\\candidate} & \shortstack{1 layer,\\fixed before\\the search} & \shortstack{4 layers,\\fixed before\\the search} & \shortstack{Selected by\\$\AULC(T)$ of\\Random alone} & \shortstack{Selected by a\\search done\\before this\\selection\\procedure\\was fixed}\\
    \midrule
    Width & Log & 16--1024 & 512 & 512 & 64 & 64 & 16 & 4\\
    \shortstack[l]{Number of\\hidden layers} & Linear & 1--8 & 2 & 2 & 1 & 4 & 7 & 2\\
    Latent dimension & Linear & 2--8 & 2 & 2 & 2 & 2 & 4 & 2\\
    Learning rate & Log & $[10^{-4},10^{-2}]$ & $5.8\times10^{-4}$ & $6.1\times10^{-4}$ & $1.1\times10^{-3}$ & $1.1\times10^{-3}$ & $1.95\times10^{-3}$ & $3.0\times10^{-3}$\\
    Adam updates & Log & 400--6400 & 5493 & 1263 & 1600 & 1600 & 6389 & 400\\
    \shortstack[l]{$L_2$ regularization\\coefficient} & Log & $[10^{-6},10^{-2}]$ & $3.5\times10^{-5}$ & $6.7\times10^{-5}$ & $5.5\times10^{-4}$ & $1.0\times10^{-6}$ & $10^{-2}$ & $1.0\times10^{-3}$\\
    Initialization & -- & default, He & default & default & default & default & He & default\\
    \midrule
    Search $R$ & & & $0.758$ & $0.750$ & $0.796$ & $0.798$ & -- & --\\
    Confirmation $R$ & & & $\mathbf{0.854}$ & $0.855$ & $0.920$ & $0.922$ & -- & $0.909$\\
    \bottomrule
  \end{tabular}
\end{table}

\paragraph{Exclusion criterion}
\label{sec:acq-preregistered}
Before the runs of this experiment, we fixed an exclusion criterion different from that of \Cref{sec:gpm-exclusion}.
A true function is excluded from the results of both OFML-GPN and OFML-Ens if any run of OFML-GPN with the four acquisition functions meets any of the following four conditions.
The condition values were recorded at each of the 93 time points with 8 to 100 observations in total, before the prediction at that time point was computed.
\begin{enumerate}
  \item The computed effective sample size is NaN or infinite.
  \item The effective sample size divided by the number of particles is below $0.10$.
  \item The pCN acceptance rate is below $0.05$ at 10 consecutive recorded time points. The acceptance rate is that of the last 8 of the 96 sweeps in the most recent assimilation with pCN sweeps, averaged over the type blocks.
Time points before the first sweep are not counted.
  \item When the marginal likelihood of the terminal type $\mathsf Q$ is computed by a low-rank approximation, the maximum over particles of the absolute difference between the low-rank and exact log-likelihoods exceeds $10^{-3}$.
\end{enumerate}
The recorded effective sample size is the value after the assimilation.
It equals the number of particles after resampling and is at least $0.65$ times the number of particles without resampling.
Therefore, condition 2 could not be met in this experiment.
In addition, the low-rank approximation is used only with 120 or more observations, and this experiment had at most 100 observations, so condition 4 could not be met either.
No true function met this criterion.

\subsection{Additional Results}
\label{sec:acq-results}

\paragraph{Differences between acquisition functions}
\label{sec:acq-paired}
\Cref{tab:acq-paired} shows the mean and 95\% interval of the per-function difference in $\AULC(T)$ over the 16 true functions.
We first consider the contribution of using predictive uncertainty.
For both surrogate models, the differences of IVR and MaxVar from Space-filling and from Random were negative, with 95\% intervals excluding zero.
That is, the contribution of using predictive uncertainty appeared not only for OFML-GPN, which satisfies the assumptions of \Cref{thm:compact-ofml}, but also for OFML-Ens, which is not correctly specified.
We next compare the two acquisition functions that do not use uncertainty.
For Random minus Space-filling, the 95\% interval included zero for OFML-GPN, and the difference was positive with a 95\% interval excluding zero for OFML-Ens.
This result shows that, for OFML-Ens, spreading the observations evenly over the candidate points itself reduced the error.
Therefore, for OFML-Ens, the contribution of using predictive uncertainty appears in the difference from Space-filling, and the difference from Random also contains this spreading effect.

\begin{table}[t]
  \centering
  \caption{Per-function differences in $\AULC(T)$ between the acquisition functions in \Cref{subsec:exp-acquisition}. The differences were computed on the 16 true functions.
  Each difference compares two acquisition functions for the same surrogate model.
  It is the $\AULC(T)$ of the acquisition function written first minus the $\AULC(T)$ of the acquisition function written second.
  The 95\% interval is the percentile bootstrap interval of the mean difference over true functions.
  Count is the number of the 16 true functions on which the acquisition function written first has the smaller $\AULC(T)$.}
  \label{tab:acq-paired}
  \small
  \input{generated_exp/table_obj2_paired_en.tex}
\end{table}

\paragraph{Coverage and calibration of the posterior variance}
\label{sec:acq-uniform}
The conclusion of \Cref{thm:compact-ofml} is that both the maximum posterior variance and the maximum posterior-mean error approach 0.
For a correctly specified surrogate model, the posterior variance at each point equals the posterior expected squared error of the posterior mean.
Therefore, reducing the posterior variance by IVR also reduces the error.
This paragraph examines whether the posterior variance represented the size of the error for the two surrogate models.

The true functions were drawn from the prior, so the expected 95\% coverage of OFML-GPN is $0.95$ if its inference is exact.
In the means over the 16 functions in \Cref{fig:exp-acquisition}(b), the 95\% coverage of OFML-GPN ranged from $0.910$ to $0.945$ on both the protocols with an initial point and those without one, and that of OFML-Ens ranged from $0.07$ to $0.26$.
That is, the coverage of OFML-GPN was close to the nominal level, and the 95\% intervals of OFML-Ens did not contain the true response at most candidate points.

We further computed two quantities for each true function and each number of observations in the runs where IVR selected the observations.
The first is Spearman's rank correlation between the posterior standard deviation $s_t(x):=\sqrt{s_t^2(x)}$ and the absolute error $\abs{\mu_t(x)-f(x)}$ over the 1024 candidate points.
It measures whether the posterior standard deviation orders the points by the size of the error.
The second is the median of the ratio $\abs{\mu_t(x)-f(x)}/s_t(x)$ over the same candidate points.
If $f(x)$ given the observations follows the normal distribution with mean $\mu_t(x)$ and variance $s_t^2(x)$, the ratio at each point follows the distribution of $\abs{Z}$ for $Z\sim\mathcal N(0,1)$, whose median is $0.674$.
For OFML-Ens, both quantities use $\tilde s_t$ instead of $s_t$.
As \Cref{tab:acq-sd-err} shows, the median ratio of OFML-Ens far exceeded $0.674$ at every number of observations.
Therefore, the posterior standard deviation of OFML-Ens greatly underestimated the size of the error.
Moreover, the rank correlation of OFML-Ens decreased from $0.34$ to $0.11$ as the number of observations increased.
That is, as observations accumulated, the posterior standard deviation of OFML-Ens corresponded less to the order of the error sizes.
Hence, the main-text observation that the maximum posterior variance of OFML-Ens stayed small while its maximum posterior-mean error remained at least $0.96$ is due to the posterior variance not representing the size of the error.
That is, for OFML-Ens, the correspondence between the posterior variance and the error that underlies the conclusion of \Cref{thm:compact-ofml} did not hold.

\begin{table}[t]
  \centering
  \caption{Spearman's rank correlation between the posterior standard deviation and the absolute error, and the median ratio of the absolute error to the posterior standard deviation, when IVR selects the observations.
  Values are means over the 16 true functions of the per-function values.
  The parentheses give the minimum and maximum over the 16 functions.}
  \label{tab:acq-sd-err}
  \small
  \begin{tabular}{@{}l r rr@{}}
    \toprule
    Surrogate model & Number of observations & Rank correlation & Median of the ratio \\
    \midrule
    OFML-GPN & 8 & $0.36$ $(0.12, 0.62)$ & $0.82$ $(0.42, 1.17)$ \\
     & 50 & $0.47$ $(0.28, 0.60)$ & $0.58$ $(0.40, 0.78)$ \\
     & 100 & $0.24$ $(0.06, 0.47)$ & $0.70$ $(0.49, 0.97)$ \\
    \addlinespace
    OFML-Ens & 8 & $0.34$ $(-0.05, 0.52)$ & $9.22$ $(3.28, 19.22)$ \\
     & 50 & $0.20$ $(-0.05, 0.39)$ & $9.61$ $(1.93, 18.84)$ \\
     & 100 & $0.11$ $(-0.09, 0.39)$ & $5.56$ $(3.27, 9.26)$ \\
    \bottomrule
  \end{tabular}
\end{table}

\paragraph{Differences between OFML-Ens hyperparameter configurations}
\label{sec:acq-ens-settings}
To examine how much the results of OFML-Ens depend on the hyperparameters, we also ran OFML-Ens with three other configurations on the 16 functions used for the results.
These runs were in addition to those with the configuration selected by the mean $\AULC(T)$ over the four acquisition functions.
The three configurations are the one selected by the $\AULC(T)$ of Random alone, the 1-layer configuration fixed before the search, and the one selected by a search done before the selection procedure of \Cref{sec:acq-ens-selection} was fixed.
\Cref{tab:acq-ens-selection} gives all of them.
As \Cref{tab:acq-ens-settings} shows, the difference of the $\AULC(T)$ of IVR minus that of Space-filling was negative for the configurations in the first and third rows and positive for those in the second and fourth rows.
That is, the result in the main text that, with OFML-Ens, the acquisition functions using predictive uncertainty reduce the error faster than Space-filling depends on the hyperparameters.
\Cref{thm:compact-ofml} gives its guarantee for correctly specified surrogate models, and this dependence for OFML-Ens, which is not correctly specified, is consistent with the scope of the guarantee.
In contrast, for all four configurations, the $\AULC(T)$ of IVR was larger than $0.479$, the value of OFML-GPN with IVR.
Therefore, OFML-GPN reduced the error faster than OFML-Ens with each of the four configurations examined.

\begin{table}[t]
  \centering
  \caption{Mean $\AULC(T)$ over the 16 true functions used for the results for each hyperparameter configuration of OFML-Ens. The Mean column is the mean over the four acquisition functions.
  The last column gives the $\AULC(T)$ of IVR minus that of Space-filling and its 95\% interval.
  Bold marks the smallest value in each column for OFML-Ens. The last row gives the values of OFML-GPN, which uses the true hyperparameters.}
  \label{tab:acq-ens-settings}
  \footnotesize
  \setlength{\tabcolsep}{3pt}
  \begin{tabular}{@{}>{\raggedright\arraybackslash}p{0.27\textwidth} rr rrrr r r@{}}
    \toprule
    Hyperparameter configuration & Width & \shortstack{Hidden\\layers} & IVR & MaxVar & \shortstack{Space-\\filling} & Random & Mean & IVR $-$ Space-filling\\
    \midrule
    Selected by the mean $\AULC(T)$ over the four acquisition functions (main text) & 512 & 2 & $0.737$ & $0.739$ & $0.809$ & $0.867$ & $\mathbf{0.788}$ & $-0.072$ $[-0.125,-0.022]$\\
    Selected by the $\AULC(T)$ of Random alone & 16 & 7 & $0.931$ & $0.968$ & $0.793$ & $0.901$ & $0.898$ & $+0.137$ $[+0.055,+0.217]$\\
    1 layer, fixed before the search & 64 & 1 & $\mathbf{0.699}$ & $\mathbf{0.699}$ & $0.877$ & $0.967$ & $0.810$ & $-0.178$ $[-0.213,-0.144]$\\
    Selected by a search done before the selection procedure of \Cref{sec:acq-ens-selection} was fixed & 4 & 2 & $0.898$ & $0.919$ & $\mathbf{0.779}$ & $\mathbf{0.847}$ & $0.861$ & $+0.119$ $[+0.076,+0.168]$\\
    \midrule
    OFML-GPN & -- & -- & $0.479$ & $0.546$ & $0.645$ & $0.664$ & $0.584$ & $-0.166$ $[-0.206,-0.135]$\\
    \bottomrule
  \end{tabular}
\end{table}

\section{Computing Resources and Software}
\label{sec:repro}

This appendix reports the computers, software, run times, and total computation of the experiments.
The experimental settings are described in \Cref{sec:exp-gpmatched,sec:acq-details}.

In \Cref{subsec:exp-transfer}, a run is the computation for one surrogate model and one true function from the source data $\mathcal D_0$ through $T=200$ added observations.
\Cref{sec:acq-details} defines a run in \Cref{subsec:exp-acquisition}.

We used the two computers listed in \Cref{tab:repro-computers}.

\begin{table}[!ht]
\centering
\small
\setlength{\tabcolsep}{3pt}
\caption{Computers used for the experiments.}
\label{tab:repro-computers}
\begin{tabular}{@{}llll@{}}
\toprule
Computer & CPUs & GPUs & Memory per GPU \\
\midrule
V100 & 2 Intel Xeon Gold 6130 & 8 NVIDIA Tesla V100 SXM2 & 16 GB \\
H100 NVL & 2 Intel Xeon Gold 6530 & 2 NVIDIA H100 NVL & 95,830 MiB \\
\bottomrule
\end{tabular}
\end{table}

Each OFML-GPN or Independent-GPN run used one GPU.
All deep-ensemble runs in \Cref{subsec:exp-transfer,subsec:exp-acquisition} used one CPU thread of the H100 NVL computer.
The runs of OFML-Ens in \Cref{subsec:exp-acquisition} flushed subnormal numbers to zero through a PyTorch setting.

Inference for OFML-GPN and Independent-GPN used JAX 0.10.2 on GPUs.
The deep-ensemble surrogate models used PyTorch 2.13.0.
We also used Python 3.12, NumPy 2.5, SciPy 1.18, and scikit-learn 1.9.

\Cref{tab:repro-runtimes} reports the median wall-clock time per run.
The run counts refer to the 12 true functions used for the results in \Cref{subsec:exp-transfer} and the runs in \Cref{subsec:exp-acquisition}.

\begin{table}[!ht]
\centering
\small
\setlength{\tabcolsep}{3pt}
\caption{Numbers of runs and median wall-clock time of one run on each computer. The deep-ensemble surrogate models ran on one CPU thread of the H100 NVL computer.}
\label{tab:repro-runtimes}
\begin{tabular}{@{}llrrrlll@{}}
\toprule
Experiment & Surrogate model
& \multicolumn{3}{c}{Number of runs}
& \multicolumn{3}{c}{Median wall-clock time} \\
\cmidrule(lr){3-5}
\cmidrule(l){6-8}
& & Total & V100 & H100 NVL
& All runs & V100 & H100 NVL \\
\midrule
\Cref{subsec:exp-transfer} & OFML-GPN
& 12 & 8 & 4 & 15.1 h & 16.8 h & 8.2 h \\
\Cref{subsec:exp-transfer} & Independent-GPN
& 12 & 8 & 4 & 3.9 h & 4.5 h & 1.7 h \\
\Cref{subsec:exp-transfer} & OFML-Ens
& 12 & 0 & 12 & 13.7 min & --- & 13.7 min \\
\Cref{subsec:exp-transfer} & Shared-DAG-Ens
& 12 & 0 & 12 & 40.7 min & --- & 40.7 min \\
\Cref{subsec:exp-transfer} & DAGNN-Cat-Ens
& 12 & 0 & 12 & 33.8 min & --- & 33.8 min \\
\Cref{subsec:exp-transfer} & Shared-Ens
& 12 & 0 & 12 & 9.5 min & --- & 9.5 min \\
\midrule
\Cref{subsec:exp-acquisition} & OFML-GPN
& 64 & 20 & 44 & 28.7 min & 80.0 min & 24.8 min \\
\Cref{subsec:exp-acquisition} & OFML-Ens
& 64 & 0 & 64 & 13.5 h & --- & 13.5 h \\
\bottomrule
\end{tabular}
\end{table}

For OFML-GPN in \Cref{subsec:exp-transfer}, the run time included the assimilation of the 80 points of $\mathcal D_0$.
The median time of this assimilation was 3.9 h.
Each deep-ensemble run in \Cref{subsec:exp-transfer} trained all five members 201 times.

For OFML-GPN in \Cref{subsec:exp-acquisition}, the run time excluded assimilation of the 8 initial points.
This assimilation was computed once per true function.
Its median time was 214 s on the V100 computer and 70 s on the H100 NVL computer.
Median run times by acquisition function ranged from 23.5 min for Random to 30.6 min for IVR.

For OFML-Ens in \Cref{subsec:exp-acquisition}, the run time included the first training on the initial points alone.
The median time of this training was 443 s.
Median run times by acquisition function ranged from 13.49 h to 13.54 h.

\Cref{tab:repro-total} reports the total recorded computation.
GPU hours were calculated by summing the elapsed times of GPU jobs, with one GPU per job.
CPU thread-hours were calculated by summing the run times in the run records, with one thread per run.
Dashes indicate inapplicable entries.

\begin{table}[!ht]
\centering
\small
\setlength{\tabcolsep}{4pt}
\caption{Recorded GPU hours and CPU thread-hours by computation. The totals were computed before rounding.}
\label{tab:repro-total}
\begin{tabular}{@{}p{0.58\linewidth}rrr@{}}
\toprule
Computation & \multicolumn{2}{c}{GPU hours}
& \shortstack{CPU\\thread-hours} \\
\cmidrule(lr){2-3}
& V100 & H100 NVL & \\
\midrule
\Cref{subsec:exp-transfer}, GP-surrogate runs on the 12 functions used for the results
& 178.5 & 44.0 & --- \\
\Cref{subsec:exp-transfer}, GP-surrogate runs on the excluded functions and on the 3 functions drawn after the fifth added function
& 255.6 & 51.4 & --- \\
\Cref{subsec:exp-transfer}, tests with modified inference settings and diagnostic runs
& 160.2 & 19.2 & --- \\
\Cref{subsec:exp-acquisition}, OFML-GPN runs and assimilation of the initial points
& 25.8 & 18.8 & --- \\
\Cref{sec:acq-details}, search and selection of the OFML-Ens hyperparameters on GPU
& 0 & 101.5 & --- \\
Development, tests, and experiments not reported in this paper
& 100.3 & 45.7 & --- \\
\midrule
\Cref{subsec:exp-transfer}, deep-ensemble runs on the 12 functions used for the results
& --- & --- & 16.7 \\
\Cref{subsec:exp-transfer}, deep-ensemble runs on the other 12 functions that were run
& --- & --- & 19.3 \\
\Cref{subsec:exp-transfer}, hyperparameter search of the four deep-ensemble surrogate models
& --- & --- & 1414.7 \\
\Cref{subsec:exp-acquisition}, OFML-Ens runs
& --- & --- & 862.5 \\
\Cref{sec:acq-details}, recomputation of the OFML-Ens candidates on the confirmation functions
& --- & --- & 3604.6 \\
Development and experiments not reported in this paper
& --- & --- & 1546.2 \\
\midrule
Total & 720.4 & 280.7 & 7463.9 \\
\bottomrule
\end{tabular}
\end{table}

CPU time used by GPU jobs was not recorded.
Times were also not recorded for two kinds of computation.
These were the trajectories partially computed in one cancelled job and the CPU jobs without per-run records.

We release the code and the run logs at \url{https://github.com/ozatamago/ofml-release}.
The run logs include the decision of the exclusion criterion for each true function.

%% file: generated_exp/table_obj1_paired_en.tex
\begin{tabular}{@{}ll rrr rrr rrr@{}}
\toprule
 & & \multicolumn{3}{c}{$\NRMSE(0)$} & \multicolumn{3}{c}{$\NRMSE(T)$} & \multicolumn{3}{c}{$\AULC(T)$}\\
\cmidrule(lr){3-5}\cmidrule(lr){6-8}\cmidrule(l){9-11}
Pair & Group & \twoline{Mean}{difference} & \twoline{95\%}{interval} & Count & \twoline{Mean}{difference} & \twoline{95\%}{interval} & Count & \twoline{Mean}{difference} & \twoline{95\%}{interval} & Count\\
\midrule
OFML-GPN $-$ & All & $-1.29$ & $[-2.14,-0.60]$ & 11/12 & $+0.01$ & $[-0.03,+0.05]$ & 7/12 & $-0.12$ & $[-0.17,-0.07]$ & 10/12 \\
Independent-GPN & Serial & $-2.35$ & $[-3.82,-1.17]$ & 12/12 & $+0.04$ & $[-0.04,+0.13]$ & 7/12 & $-0.14$ & $[-0.24,-0.03]$ & 10/12 \\
 & Merge & $-1.11$ & $[-1.83,-0.48]$ & 11/12 & $0.00$ & $[-0.04,+0.03]$ & 7/12 & $-0.14$ & $[-0.19,-0.08]$ & 11/12 \\
\addlinespace
OFML-GPN $-$ & All & $-0.14$ & $[-0.24,-0.04]$ & 9/12 & $-0.10$ & $[-0.15,-0.06]$ & 11/12 & $-0.17$ & $[-0.21,-0.14]$ & 12/12 \\
OFML-Ens & Serial & $-0.26$ & $[-0.42,-0.11]$ & 9/12 & $-0.09$ & $[-0.17,-0.01]$ & 9/12 & $-0.14$ & $[-0.22,-0.05]$ & 10/12 \\
 & Merge & $-0.13$ & $[-0.26,-0.01]$ & 8/12 & $-0.14$ & $[-0.22,-0.08]$ & 11/12 & $-0.24$ & $[-0.29,-0.20]$ & 12/12 \\
\addlinespace
Shared-DAG-Ens $-$ & All & $-0.42$ & $[-0.70,-0.15]$ & 8/12 & $-0.42$ & $[-0.52,-0.31]$ & 12/12 & $-0.37$ & $[-0.48,-0.26]$ & 12/12 \\
Shared-Ens & Serial & $-0.31$ & $[-0.48,-0.16]$ & 10/12 & $-0.38$ & $[-0.49,-0.25]$ & 11/12 & $-0.35$ & $[-0.45,-0.25]$ & 12/12 \\
 & Merge & $-0.45$ & $[-0.89,-0.02]$ & 8/12 & $-0.45$ & $[-0.56,-0.34]$ & 12/12 & $-0.39$ & $[-0.50,-0.28]$ & 12/12 \\
\addlinespace
OFML-Ens $-$ & All & $+0.02$ & $[-0.04,+0.09]$ & 4/12 & $+0.01$ & $[-0.02,+0.04]$ & 4/12 & $+0.04$ & $[+0.02,+0.05]$ & 1/12 \\
Shared-DAG-Ens & Serial & $+0.00$ & $[-0.09,+0.10]$ & 7/12 & $+0.00$ & $[-0.05,+0.04]$ & 6/12 & $+0.02$ & $[-0.02,+0.05]$ & 5/12 \\
 & Merge & $-0.01$ & $[-0.16,+0.13]$ & 5/12 & $+0.02$ & $[-0.02,+0.06]$ & 4/12 & $+0.06$ & $[+0.04,+0.07]$ & 0/12 \\
\addlinespace
OFML-Ens $-$ & All & $-0.10$ & $[-0.19,-0.03]$ & 9/12 & $-0.01$ & $[-0.04,+0.01]$ & 9/12 & $+0.00$ & $[-0.02,+0.02]$ & 6/12 \\
DAGNN-Cat-Ens & Serial & $-0.20$ & $[-0.35,-0.04]$ & 9/12 & $-0.02$ & $[-0.04,+0.01]$ & 6/12 & $-0.04$ & $[-0.08,-0.01]$ & 9/12 \\
 & Merge & $-0.11$ & $[-0.26,+0.04]$ & 7/12 & $-0.01$ & $[-0.06,+0.04]$ & 7/12 & $+0.03$ & $[0.00,+0.07]$ & 3/12 \\
\bottomrule
\end{tabular}

%% file: generated_exp/table_obj2_paired_en.tex
\begin{tabular}{@{}l rrr rrr@{}}
\toprule
 & \multicolumn{3}{c}{OFML-GPN} & \multicolumn{3}{c}{OFML-Ens}\\
\cmidrule(lr){2-4}\cmidrule(l){5-7}
Pair & \twoline{Mean}{difference} & \twoline{95\%}{interval} & Count & \twoline{Mean}{difference} & \twoline{95\%}{interval} & Count\\
\midrule
IVR $-$ Space-filling & $-0.166$ & $[-0.206,-0.135]$ & 16/16 & $-0.072$ & $[-0.125,-0.022]$ & 13/16 \\
MaxVar $-$ Space-filling & $-0.099$ & $[-0.147,-0.048]$ & 14/16 & $-0.070$ & $[-0.123,-0.017]$ & 13/16 \\
Random $-$ Space-filling & $+0.019$ & $[-0.016,+0.057]$ & 8/16 & $+0.058$ & $[+0.017,+0.099]$ & 6/16 \\
IVR $-$ Random & $-0.185$ & $[-0.221,-0.153]$ & 16/16 & $-0.130$ & $[-0.198,-0.071]$ & 15/16 \\
MaxVar $-$ Random & $-0.119$ & $[-0.174,-0.059]$ & 14/16 & $-0.128$ & $[-0.198,-0.066]$ & 13/16 \\
\bottomrule
\end{tabular}